\newcommand{\arxivversion}{}
\documentclass{article}

\PassOptionsToPackage{numbers, compress}{natbib}
\usepackage[final,main]{neurips26/neurips_2026}

\usepackage[utf8]{inputenc} % allow utf-8 input
\usepackage[T1]{fontenc}    % use 8-bit T1 fonts
\usepackage[hidelinks]{hyperref} % clickable links without printed borders
\usepackage{url}            % simple URL typesetting
\usepackage{booktabs}       % professional-quality tables
\usepackage{multirow}       % multi-row table cells
\usepackage{amsmath}        % math environments and \text
\usepackage{amsfonts}       % blackboard math symbols
\usepackage{nicefrac}       % compact symbols for 1/2, etc.
\usepackage{microtype}      % microtypography
\usepackage{xcolor}         % colors
\usepackage{graphicx}       % include graphics
\usepackage{subcaption}     % subfigures with separate captions/labels
\usepackage{wrapfig}        % figures beside text
\usepackage[most]{tcolorbox} % styled boxes for prompts
\usepackage{etoc}           % local table of contents for appendix
\usepackage{xspace}         % smart trailing space after text macros

\newcommand{\Sset}{\mathcal{S}}           % skill space
\newcommand{\Dset}{\mathcal{D}}           % demonstration set
\newcommand{\sprior}{s_{\text{prior}}}     % prior skill
\newcommand{\sground}{s_{\text{ground}}}   % grounding skill

\newcommand{\prior}{training prior\xspace}     % singular, body text
\newcommand{\priors}{training priors\xspace}   % plural, body text
\newcommand{\Prior}{Training Prior\xspace}     % title case (paper title, section headings)

\newcommand{\compositions}{compositions\xspace}   % plural, body text
\newcommand{\setting}{setting\xspace}             % singular, body text
\newcommand{\settings}{settings\xspace}           % plural, body text
\newcommand{\Setting}{Setting\xspace}             % title case
\newcommand{\Settings}{Settings\xspace}           % title case plural

\graphicspath{{fig/}}

\title{In-Context Learning Can Help Vision-Language Models Overcome Training Priors}
\author{%
  Kun Wang \quad
  Xindi Wu \quad
  Sanghyuk Chun \quad
  Olga Russakovsky \quad
  Esin Tureci\\
  Princeton University \\
  \texttt{\{bw5889, xw5676, sanghyukc, olgarus, eturci\}@princeton.edu}
}

\begin{document}

\maketitle

% Abstract
\begin{abstract}
Vision-language models (VLMs) learn strong statistical regularities during training, which can make them fail to properly respond to visual evidence when input images violate those regularities. It is challenging to disambiguate these prior-induced errors from errors caused by the model missing the underlying visual capabilities. In this work, we use controlled 
visual in-context learning to  encourage the model to focus on the visual evidence, and demonstrate that this alone is frequently sufficient to correct the errors -- suggesting that the models in fact do possess the relevant visual capabilities. 
Specifically, we show that across five vision-centric tasks, demonstrations yield large gains on prior-conflicting counterfactual examples ($+27.2\%$ for Qwen3-VL 235B and $+15.1\%$ for Qwen3-VL 32B) while leaving accuracy on prior-aligned examples nearly unchanged. These findings  generalize across five VLM families with gains from $+9.7\%$ to $+30.3\%$.
We analyze visual representations and find that demonstrations indeed redirect attention toward grounding cues while reducing reliance on cues that support the learned prior. Finally, we find that in-context learning is limited when the underlying visual capability is weak or absent. 
% \olga{I think this previous sentence either needs to be explained (ok, but how do we know that the visual capability is actually weak or absent?) or simply removed (that would be my suggestion)}
Together, these results show that current VLMs possess visual grounding capabilities that standard evaluations may overlook, underscoring the need for controlled evaluation design in diagnosing and improving model capability.
Code: \url{https://github.com/princetonvisualai/vlm_icl_tp}.
% \xindi{nitpicking: can we shorten the current title a bit? :p I generally like short \& powerful titles but no worries if not:)
% }
% \kun{TO BE REVISED}
% Vision-language models often fail on visual tasks when the correct answer conflicts with common training priors. 
% Such errors are often treated as missing visual capability and addresses them by changing training, model internals, or input preprocessing. 
% We argue instead that the model has the needed visual evidence and abilities, but its answer is dominated by a learned prior. 
% To test this, we use visual in-context learning as a diagnostic, as it changes only the context while keeping model weights, the visual encoder, and image preprocessing fixed. 
% We build a controlled benchmark with prior-consistent and prior-conflicting examples paired with label-balanced demonstrations. 
% In-context learning improves performance most on prior-conflicting examples. 
% Mechanistic analysis on attention finds that demonstrations reduce reliance on prior cues and increase reliance on visual evidence. 
% Further experiments show that in-context learning can reveal capabilities suppressed by priors, but cannot create visual capabilities that are absent. 
% Our work shows that some VLM failures reflect learned statistical priors rather than missing visual capabilities. 
% We introduce controlled settings that reveal when in-context learning can help VLMs use those skills, offering a useful framework for studying continual improvement in VLMs.

\end{abstract}

% Main sections
% Introduction

\section{Introduction}
\begin{figure}[t]
    \centering
    \includegraphics[width=\linewidth]{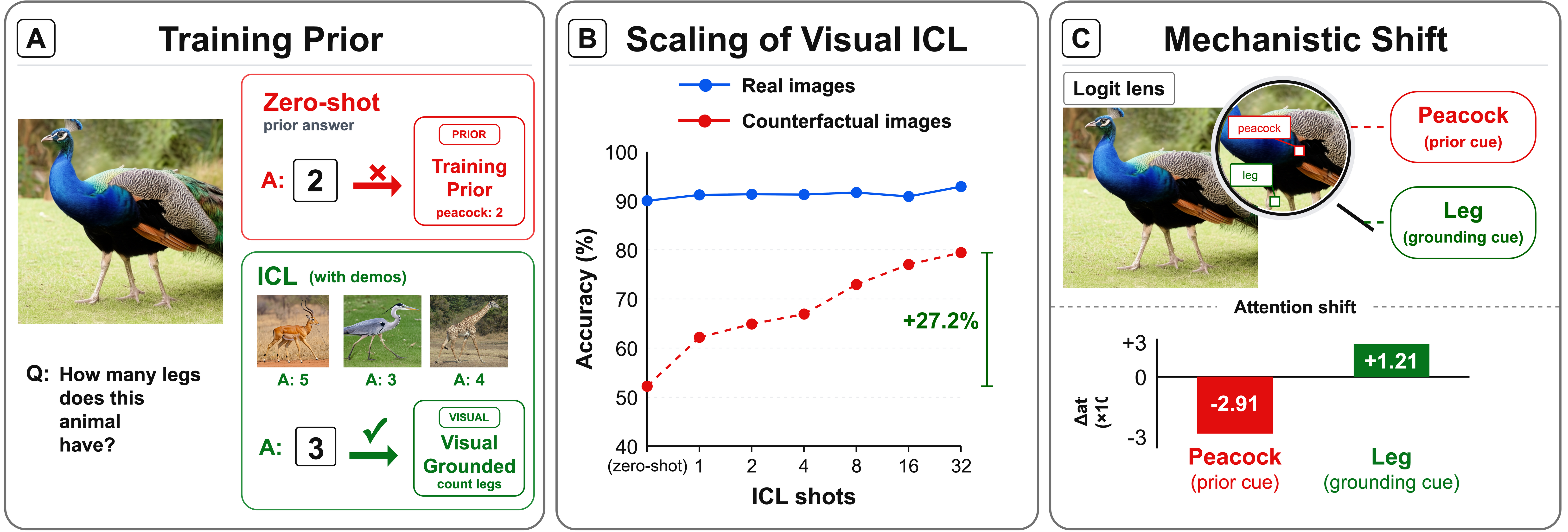}
    \caption{\textbf{Visual ICL overcomes \prior dominance.} \textbf{(A)} A VLM answers ``2'' for a three-legged peacock under zero-shot prompting; label-balanced demonstrations shift the answer to the visually grounded ``3.'' \textbf{(B)} Scaling ICL shots yields a $+27.2$ pp accuracy gain on counterfactual images while real images remain near ceiling. \textbf{(C)} Mechanistically, demonstrations redirect attention from the prior cue (``peacock'') to the grounding cue (``leg'').}
    \label{fig:1_diagram}
\end{figure}

In humans, visual intelligence necessitates grounding abstract knowledge and reasoning directly within the perceived image. 
Yet, vision-language models (VLMs) frequently falter when an image departs from the dominant patterns of their training data~\citep{luo2025probingvisuallanguagepriors,vo2025visionlanguagemodelsbiased,tong2024eyeswideshutexploring,asadi2026mirageillusionvisualunderstanding}:
when shown an image depicting a three-legged peacock and asked how many legs are visible, models often answer ``2'' as shown in Figure \ref{fig:1_diagram}(a). %Figure \ref{fig:training_prior} (b). 
Existing work addresses these failures by intervening mechanistically~\citep{kang2025yourground,zhao2025tellground,wan2024contrastiveground,li2026makecama}, changing training procedures~\citep{qwen2025qwen25technicalreport,bai2025qwen3vltechnicalreport,meng2024deepstack,gemmateam2025gemma3technicalreport}, or improving visual input preprocessing~\citep{wu2024v,shen2025zoomeye,shtedritski2023doesredcircle}.

% These approaches rests on the same assumption: that the model either lacks the relevant visual information or cannot incorporate it on its own.

% These approaches treat visual grounding failures as evidence that the model lacks access to the relevant visual representations. ("aim", capabilities)
These approaches aim to address visual grounding failures by providing models with stronger visual signals and capabilities, treating such failures as evidence that the necessary capabilities are missing.
Recent work, however, shows that the relevant visual evidence is already encoded in the internal representations of VLMs~\citep{fu2025hiddenplainsightvlms,liu2025visualrepresentationsinsidelanguage}.
This leaves an open question: are these failures caused by the model's inability to incorporate visual information, or by a stronger learned expectation that overrides the accessible grounding signal? 
%\esin{Also here, to make it consistent with the next paragraph rephrase as which of these failures reflect ...} 
% In-context learning (ICL) provides a natural test: demonstrations can steer a model without changing its parameters \citep{brown2020languagemodelsfewshotlearners}, allowing us to ask whether errors persist when the context emphasizes visual evidence in the image over learned expectations.

In-context learning (ICL) provides a natural test: providing demonstrations of other animals with answers (e.g., a five-legged dog, a two-legged parrot) informs the model that the task is to count the legs visible in the image~\citep{brown2020languagemodelsfewshotlearners}.
If errors persist despite such demonstrations, the model likely lacks the visual capability to count legs in the first place.  
If demonstrations correct the error, the model can already count legs but its learned knowledge about the animal overrides the visual evidence.

Although ICL is well established as a steering mechanism in language models \citep{wei2023largerlanguagemodelsincontext,pawelczyk2024incontextunlearninglanguagemodels,pan2023incontextlearninglearnsincontext,kossen2024incontextlearninglearnslabel,lampinen2025generalizationlanguagemodelsincontext}, 
prior work has reported that visual ICL lags behind its language counterpart \citep{zong2025vliclbenchdevildetails,wang2026multimodalincontextlearninglags}.
These studies have made important progress, but they leave open the question of whether demonstrations can shift a model toward visually grounded answers.
Many existing studies test surface-level image--label associations, such as mapping pictures of lions to the label ``dog'' \citep{zong2025vliclbenchdevildetails,wang2026multimodalincontextlearninglags}, where the task to be learned does not focus on visual perception and instead can be collapsed into a text-only mapping problem.
Others retrieve demonstrations from general VQA pools \citep{chen2025truemultimodalincontextlearning,baldassini2024makesmultimodalincontextlearning,huang2025mimickingreasoningrethinkingmultimodal,xu-etal-2025-introspection-analysis-icl}, where loose query--demo relevance tends to reduce ICL to format learning. 
Thus, it is still an open question whether, and for which tasks, demonstrations can actually move a model toward grounded visual evidence.
% As a result, whether demonstrations can move a model toward grounded visual evidence remains largely open.

We address this question by using ICL as a diagnostic tool to study a phenomenon we call \emph{\prior dominance}, where a model's learned statistical regularity from training overrides visual evidence during generation.
We construct a benchmark that separates prior-consistent from prior-conflicting examples and supports label-balanced demonstrations, allowing us to control the demonstration context through task-relevant examples and answer choices.
With these controls, we find that many visual grounding failures are not caused by an inability to incorporate visual information, but by overpowering \priors that models can suppress through ICL.
We further analyze this effect mechanistically, showing how demonstrations shift the models' attention toward visually grounded evidence, and we test whether demonstrations can also teach capabilities that are absent.
Together, these results suggest that addressing prior-driven failures does not require architectural modification or enhanced visual features; 
effective mitigation can instead focus on managing the influence of training priors during inference to keep the model anchored to the visual evidence.

Our main contributions are:
(1) a controlled benchmark that separates prior-consistent from prior-conflicting examples and supports label-balanced demonstrations for visual ICL;
(2) empirical and mechanistic evidence that visual ICL improves accuracy most on prior-conflicting examples, distinguishing a visual grounding shift account from simple format learning;
and (3) diagnostic analyses that test whether visual ICL teaches new visual capabilities or reveals capabilities already available but dominated by training priors.

% Our main contributions are:
% \begin{itemize}
%     \item We introduce a controlled benchmark that separates prior-consistent from prior-conflicting examples and supports label-balanced demonstrations for visual ICL.
%     \item We show that visual ICL improves accuracy most on prior-conflicting examples, distinguishing a grounding-shift account from a simple format-learning account.
%     \item We provide diagnostic and causal analyses of this shift, including token-masking experiments, and test whether visual ICL teaches new visual capabilities or reveals capabilities already available but dominated by training priors.
% \end{itemize}
%%%%%%%%%%%%%%%%%%%%%%%%%%%%%%%%%%%%%%%%%%%%%%%%%%%%%%%%%%%%%%%%%%%%%%%%%%%%%%%%%%%%%%%%%%%%%%%%%%%%%%%%

% For part that doesn't work: high-count is good setting that ICL can identify model's inherent limitations

\section{Problem Setting}
\label{sec:formal_definition}
To test whether ICL overcomes \prior dominance, we compare examples where learned expectations agree with visual evidence to examples where they conflict.

\noindent \textbf{Training Priors and Visual Grounding.}
We use \textbf{\prior} to denote a statistical regularity learned from the training distribution that biases predictions on a downstream task. 
This includes language and world-knowledge priors~\citep{luo2025probingvisuallanguagepriors,lee2025vlindbenchmeasuringlanguagepriors,golovanevsky-etal-2025-pixels-priors}, such as typical object attributes, as well as visual or spatial priors~\citep{kamath2023whatsupvisionlanguagemodels,jung2025rightrightenhancingobject}, such as common poses and layouts.

A model can answer an animal-leg question by recognizing the animal and recalling its typical leg count, or by localizing and counting its visible legs.
These compositions of capabilities agree on a typical two-legged peacock but disagree on the three-legged peacock in Figure~\ref{fig:1_diagram}(a).

\noindent \textbf{Prior-Consistent and Prior-Conflicting Examples.}
For an image--question pair $(I,Q)$, we call the image-supported answer the \emph{grounded answer}.
When a learned regularity favors an identifiable answer, we call it the \emph{prior answer}.
An example is \emph{prior-consistent} when these answers agree and \emph{prior-conflicting} when they differ.
Prior-consistent examples alone cannot distinguish a correct shortcut from visual grounding.

\noindent \textbf{Visual ICL as a Diagnostic.}
Given a query $(I_q,Q_q)$ and $k$ demonstrations $\Dset=\{(I_i,Q_i,A_i)\}_{i=1}^{k}$, visual ICL predicts an answer $A$ without changing model parameters.
We view ICL as shifting probability mass over latent compositions of capabilities~\citep{xie2021explanation,zhang2023andhowdoesincontextlearninglearn,wakayama2025incontextlearningprovablybayesian}:
\begin{align}
  p(A \mid I_q, Q_q, \Dset)
  = \sum_{s \in \Sset} p(A \mid I_q, Q_q, s)\, p(s \mid I_q, Q_q, \Dset),
\end{align}
where $s$ denotes a candidate capability composition and $\Sset$ is the space of such compositions.
We use $\sprior$ and $\sground$ to denote a prior-based composition and a visually grounded composition, respectively.
Zero-shot prompting may place high mass on $\sprior$ when the learned regularity is salient.
Demonstrations that consistently exhibit visual grounding can shift mass toward $\sground$.

This view predicts that ICL gains should concentrate on prior-conflicting examples if demonstrations help override \priors, rather than merely teach the response format.
Section~\ref{sec:setting} develops this diagnostic into a benchmark with closed-label visual tasks, label-balanced demonstrations, and controlled \settings for comparing prior-consistent and prior-conflicting examples.

% This framing differs from prior multimodal ICL evaluations that focus on format imitation or input-output mapping~\citep{chen2025truemultimodalincontextlearning,huang2025mimickingreasoningrethinkingmultimodal,li2026makecama,zong2025vliclbenchdevildetails}.
% On the other hand, prior-probing datasets show that VLMs rely on language or visual shortcuts, but they are not designed for controlled ICL evaluation~\citep{vo2025visionlanguagemodelsbiased,luo2025probingvisuallanguagepriors}.
% Our question is vision-centric: can demonstrations move a model from $\sprior$ to $\sground$ when both compositions are available? 
% The same framing also predicts a limitation: when $\sground$ is not reliable for the model, ICL should not be expected to create that capability from demonstrations alone.

\section{ICL Construction for Visual Perception and Reasoning}
% Exported from vis/3_dataset/3_vibe_design.drawio at 300% scale.
\begin{figure}[t]
    \centering
    \includegraphics[width=0.9\linewidth]{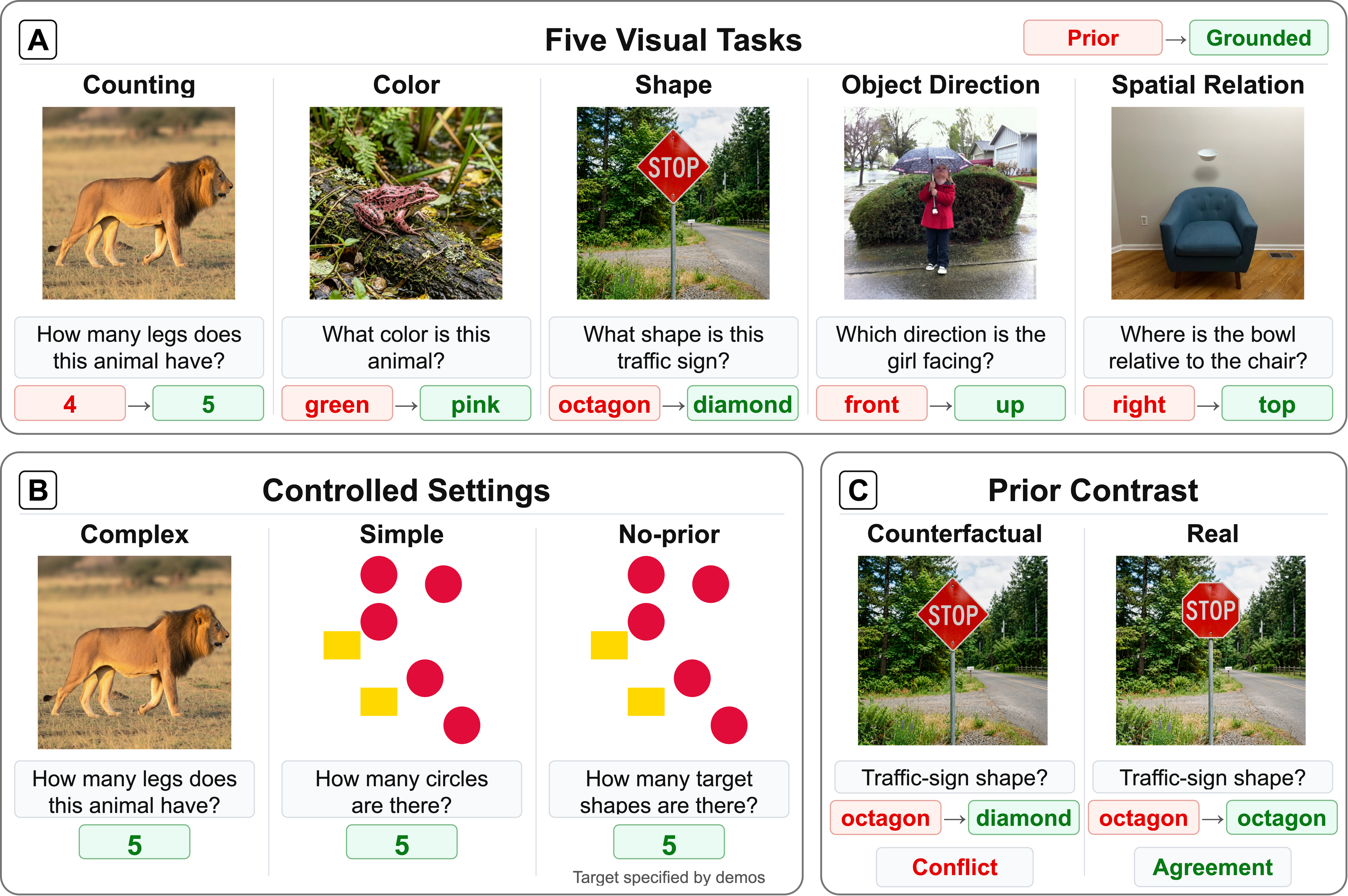}
    \caption{\textbf{Data design.} (A) Our benchmark covers five vision-centric tasks, each contrasting a prior answer with the grounded visual answer. (B) Each task contains complex, simple, and no-prior \settings to separate prior override from ordinary visual task learning. (C) We probe the training prior by comparing counterfactual examples, where the prior and visual answers conflict, with real examples, where they agree.
    % \tyler{Make it clear which is which. I want (a) to be the dataset, which means i should see what the questions are or at least some explanation as to the legend.} 
    }
    \label{fig:data_tasks}
\end{figure}

\label{sec:setting}
Prior visual ICL benchmarks fail to separate visual grounding from a learned \prior or response-format imitation \citep{huang2025mimickingreasoningrethinkingmultimodal,zong2025vliclbenchdevildetails,baldassini2024makesmultimodalincontextlearning}, while prior-probing benchmarks document \prior reliance but lack the controls needed for ICL evaluation~\citep{lee2025vlindbenchmeasuringlanguagepriors,luo2025probingvisuallanguagepriors,golovanevsky-etal-2025-pixels-priors,vo2025visionlanguagemodelsbiased}.
Our work bridges these gaps.
Each example carries a \emph{grounded answer} determined by image evidence and, when applicable, a \emph{prior answer} favored when that evidence is ignored; comparing ICL gains on \emph{prior-consistent} examples (answers agree) with \emph{prior-conflicting} examples (answers disagree) isolates \prior override from format or label learning.
To ensure this diagnostic is valid, we use label-balanced demonstrations so gains cannot be explained by frequency shortcuts (Section~\ref{sec:data_design_requirements}), difficulty-controlled \settings that isolate how the \prior affects performance (Section~\ref{sec:dataset_settings}), and vision-centric closed-label tasks where the task requires visual grounding (Section~\ref{sec:dataset_tasks}).

% 4. Data design requirements
\subsection{Data Design Requirements}
\label{sec:data_design_requirements}
The benchmark uses closed-label, vision-centric tasks with balanced answer labels and matched control settings, so the accuracy gains from $k$-shot ICL over the zero-shot baseline cannot be explained solely by response-format imitation or answer-frequency shifts. 
We construct each example as an $(I,Q,A)$ triple and use this label-balanced candidate pool for demonstration sampling.

To probe the \prior with minimal confounding factors, we use controlled counterfactual editing: we change only the queried visual property while preserving the object identity, scene context, and prompt format.
For example, Figure~\ref{fig:data_tasks} (C) edits an octagonal stop sign into a diamond-shaped sign while keeping the rest of the image fixed, making the grounded shape conflict with the canonical stop-sign prior.
To avoid a source-domain confound, prior-consistent images are also generated or regenerated with Gemini and passed through the same validation, balancing, and downsampling pipeline as the prior-conflicting images.
Two annotators independently label each complex image while blind to the intended label. 
We retain an image only if both annotators select the intended label and confirm that it meets the quality criteria.
Appendix~\ref{app:data_construction} gives the construction procedure, task-specific rejection criteria, and validation statistics.
% \olga{Is this a human study, or did you just do this in-house? If recruited folks, should provide more info about where they were recruited from, how much they were paid, etc} 

% 5. Settings
\subsection{\Settings}
\label{sec:dataset_settings}
To examine visual ICL and its ability to override the \prior, we use difficulty-controlled \settings that vary whether a prior answer is identifiable.
Each task is therefore evaluated in three \settings of increasing complexity, shown in Figure~\ref{fig:data_tasks} (B).
The \emph{simple} \setting tests the visual operation on abstract stimuli without an identifiable prior answer.
The \emph{no-prior} \setting further abstracts the prompt by replacing object names with placeholders such as ``target object,'' following the ICL view of demonstrations as task-rule evidence~\citep{xie2021explanation,min2022rethinking,pan2023incontextlearninglearnsincontext,kossen2024incontextlearninglearnslabel}; we therefore interpret its 0-to-$k$ gains as in-context task specification with language priors removed, not as an unambiguous zero-shot visual baseline.
The \emph{complex} \setting is the main evaluation: examples preserve object identity and task format while varying whether the grounded answer agrees with the prior answer.\label{sec:real_counterfactual}
In \emph{real} images the two agree; in \emph{counterfactual} images the queried visual evidence is changed so that the prior answer is wrong.
Complex examples thus test \prior override, while simple and no-prior examples control for ordinary visual competence and task-format learning.

% 6. Tasks
\subsection{Tasks}
\label{sec:dataset_tasks}

We evaluate five closed-label visual tasks: counting, color identification, shape identification, object direction, and spatial relation.
Each of these tasks supports exact-match scoring, label-balanced demonstrations, and controlled prior-conflicting examples that change only the queried visual evidence.
Together, they cover two shortcut regimes relevant to \prior dominance: counting, color identification, and shape identification test language or world-knowledge priors tied to object identity, while object direction and spatial relation test visual-configuration priors tied to common poses and layouts.
For each task, the answer label set is fixed across \settings, and accuracy is measured by exact match after task-specific normalization.
Table~\ref{tab:dataset_stats} lists sources, labels, and counts for each task and \setting, including the counting datasets used in Section~\ref{sec:limitations}.
The original complex setting contains 1,160 examples: 650 real ($56.0\%$) and 510 counterfactual ($44.0\%$).
We also evaluate an expanded version with approximately 400 complex examples per task; all additions are counterfactual (Appendix~\ref{app:expanded_benchmark}).

\noindent \textbf{Counting.} The task asks for the number of visible animal legs; the complex
setting violates category priors such as two-legged birds or four-legged
mammals \citep{vo2025visionlanguagemodelsbiased}.

\noindent \textbf{Color Identification.} The task asks for the displayed animal color, with
animals rendered in colors that conflict with their canonical real-world
coloration \citep{luo2025probingvisuallanguagepriors}.

\noindent \textbf{Shape Identification.} The task asks for the displayed traffic-sign shape,
where the sign identity remains recognizable but the rendered shape may
differ from the canonical one \citep{luo2025probingvisuallanguagepriors}.

\noindent \textbf{Object Direction.} The task asks for the camera-relative orientation (front, back,
left, right, up, down) of an object whose canonical pose acts as a shortcut; the complex
setting emphasizes rare orientations such as up and
down \citep{jung2025rightrightenhancingobject}.

\noindent \textbf{Spatial Relation.} The task asks for the relation of a target object to an
ordered reference object, where the depicted layout conflicts with the
most common configuration \citep{kamath2023whatsupvisionlanguagemodels}.

The selected source entries contain zero up/down examples for object direction and zero top/bottom examples for spatial relation.
These source-label gaps motivate the rare-configuration tests; Appendix~\ref{app:configuration_priors} reports model-side prior diagnostics, since model-specific training frequencies are unavailable.

% 8. In-context demonstrations
\subsection{In-Context Demonstrations}
Demonstrations are sampled from the same task and \setting as the query unless otherwise stated, with query images excluded from the demonstration pool.
We evaluate 0, 1, 2, 4, 8, 16, and 32 shots.
For each nonzero shot count, we pre-sample 10 fixed contexts per task and \setting and share them across queries.
We report mean accuracy and standard deviation across these contexts; each query-context pair is scored from one sampled output, so the standard deviation also includes decoding noise.
Contexts are label-balanced whenever possible, and complex-setting contexts also balance real and counterfactual examples when possible; this controls for label-frequency effects and prevents gains from being explained by changes in the marginal answer distribution.
All shot counts use the same instruction template, with only the number of demonstrations changed; Appendix~\ref{app:evaluation_settings} gives full prompt and sampling details, and Figure~\ref{fig:benchmark_prompts} lists task-specific templates.

% 9. Models and inference
\subsection{Models and Inference}

For performance analysis, we evaluate Qwen3-VL variants from 2B to 235B parameters~\citep{bai2025qwen3vltechnicalreport}.
To test generality beyond one model family, we also evaluate Qwen2.5-VL~\citep{qwen2025qwen25technicalreport}, Qwen3.5~\citep{qwen35blog}, InternVL3.5~\citep{wang2025internvl35advancingopensourcemultimodal}, and Gemma-4~\citep{gemma4_2026}.
Mechanistic experiments cover Qwen3-VL 4B and 32B and Gemma-4 E4B and 31B; masking experiments use the larger model from each family.
All models use the same prompt templates, demonstration sets, and answer normalization.
Appendix~\ref{app:model_settings} provides checkpoints and inference backends, sampling parameters, and normalization rules.

% Scaling of ICL
\section{In-Context Learning Overcomes Training Priors}
\label{sec:analysis}

Building on our diagnostic framing of ICL, this section asks when and why visual demonstrations help VLMs suppress \priors and rely on image-grounded evidence.
We ask whether demonstrations improve performance in general and on prior-conflicting examples, how they change the use of visual evidence in VLMs, and whether they can strengthen new or weak capabilities.
% we organize the analysis around three questions.
% First, do demonstrations improve performance specifically when the model's learned \prior conflicts with the image, or do they merely induce format learning?
% Second, when performance improves, do demonstrations change which visual evidence the model relies on?
% Third, can demonstrations teach weak capability compositions, or do they mainly select among capability compositions already available to the model?

% Across these analyses, visual ICL mainly resolves \prior dominance: demonstrations improve counterfactual examples, shift the model toward grounding cues, and help select capability compositions already available to the model, but they do not reliably teach weak ones.

%%%%%%%%%%% subsection 1: performance evaluation %%%%%%%%%%%
\subsection{Visual ICL Improves Performance on Prior-Conflicting Examples}
\label{sec:performance_evaluation}

We begin with the first diagnostic question: do demonstrations induce format learning, or help the model override a learned \prior when visual evidence conflicts with the prior-favored answer?
If ICL mainly induces format learning, then examples with the same prompt structure and answer space should improve similarly. 
If ICL overrides \priors, then gains should concentrate where the visually grounded answer contradicts learned regularities.

%
%\esin{Might be worth repeating what tasks we are plotting the results for in Figures 3 and 4).}

% Figure: scaling of ICL, by settings
% Main scaling figure for Qwen3VL
% Aggregated performance of Qwen3VL across different tasks for each setting
\begin{figure}[t!]
    \centering
    \includegraphics[width=\linewidth]{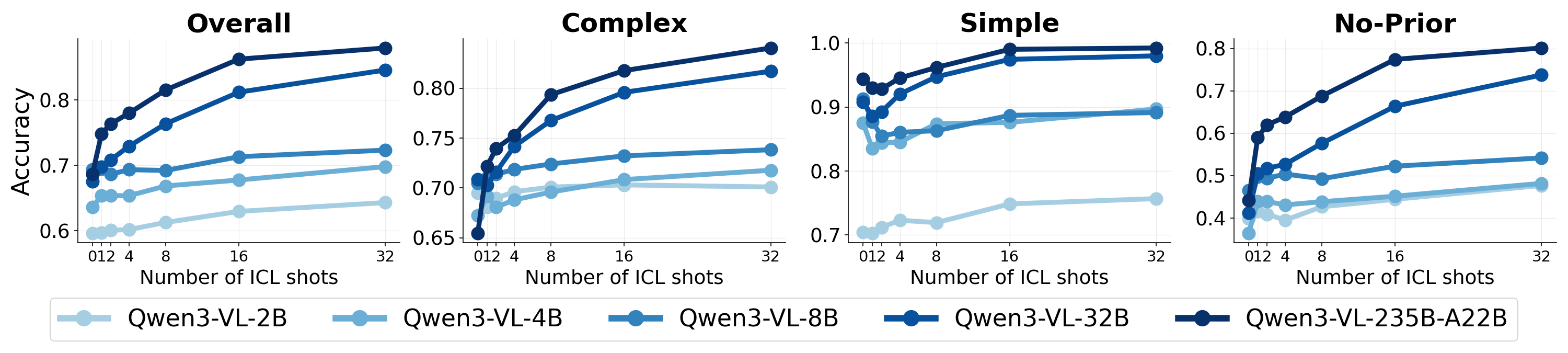}
    \caption{\textbf{Qwen3-VL scales with in-context demonstrations.} Accuracy vs.\ number of in-context demonstrations for Qwen3-VL model family on complex, simple, and no-prior \settings across tasks.}
    \label{fig:qwen3vl_scaling}
\end{figure}

\paragraph{Controlled visual ICL scales with model size and shot count.}
Figure~\ref{fig:qwen3vl_scaling} reports accuracy across shot counts for Qwen3-VL models from 2B to 235B parameters.
Accuracy increases with more demonstrations across all three \settings, and the larger models continue to gain even from 16 to 32 shots.
This trend shows that VLMs can use visual demonstrations as in-context task evidence, with stronger scaling at larger model sizes.

The setting-level comparison helps separate task and format learning from \prior override. In particular, we observe that zero-shot accuracy on \emph{simple} is already high. Given the same task format between settings, the larger gains on \emph{complex} than on \emph{simple}, especially for the 32B and 235B models, suggest that demonstrations do more than teach the response format.
They help large models perform task learning and may encourage the use of a visually grounded composition.

% The \emph{simple} and \emph{complex} settings share the same task and prompt format, but only the \emph{complex} setting introduces a prior-favored answer that can compete with the visually grounded answer.
% The \emph{no-prior} setting validates task-rule learning from demonstrations.

% Figure~\ref{fig:qwen3vl_scaling} reports accuracy across shot counts for Qwen3-VL models from 2B to 235B parameters.
% Accuracy generally increases with more demonstrations, showing that VLMs can adapt to these visual tasks in context.
% Notably, the performance gains are not uniform: they are largest in the prior-conflict examples and increase as the model's parameter size increases.
% The trend suggests that a view based only on format learning does not hold because complex and simple settings share the same prompt and answer format.
% Instead, demonstrations appear to help the model select a visually grounded composition when its default capability composition follows a misleading learned \prior.
% Consistent with this interpretation, shuffling demonstration labels preserves the input format but breaks the image-label relation, and this substantially degrades performance (Appendix~\ref{app:unlearning})~\citep{kossen2024incontextlearninglearnslabel}.%\kun{this may be optional}

% Overall accuracy on the complex setting for ~30B model families at 0/16 shots.
\begin{wraptable}{r}{0.45\textwidth}
    \footnotesize
    \setlength{\tabcolsep}{4pt}
    \renewcommand{\arraystretch}{0.95}
    \vspace{-\baselineskip}
    \centering
    \caption{In-context scaling of $\sim$30B model families on the complex setting. Accuracies and deltas are reported with \% and micro-averaged over query entries.}
    \label{tab:modelfamily_scaling}
    \begin{tabular}{lccc}
        \toprule
        \textbf{Model} & \textbf{0-shot} & \textbf{16-shot} & $\mathbf{\Delta_{16-0}}$ \\
        \midrule
        Qwen3-VL 32B    & 69.9\%          & 79.6\%          & $+9.7\%$ \\
        Qwen2.5-VL 32B  & 49.1\%          & 64.7\%          & $+15.5\%$ \\
        Qwen3.5 27B     & 44.1\%          & 74.5\%          & $\mathbf{+30.3\%}$ \\
        InternVL3.5 38B & 52.4\%          & 70.6\%          & $+18.3\%$ \\
        Gemma-4 31B     & \textbf{75.2\%} & \textbf{86.6\%} & $+11.5\%$ \\
        \bottomrule
    \end{tabular}
\end{wraptable}

To validate that this trend is not specific to Qwen3-VL, we extend the comparison to five VLMs at roughly 30B scale (Table~\ref{tab:modelfamily_scaling}).
Every family improves from 0-shot to 16-shot prompting on the aggregate complex setting, with gains ranging from $+9.7\%$ for Qwen3-VL to $+30.3\%$ for Qwen3.5.
At the per-task level (Table~\ref{tab:app_cross_family_accuracies}), complex deltas remain positive in 23 of 25 task-family cells at $\sim$30B. 
Full accuracy results across families, tasks, and settings are in Appendix~\ref{app:further_evaluations}.
% Full accuracy results across model families and tasks are in Appendix~\ref{app:further_evaluations}
% , and report results for reasoning VLMs in Appendix~\ref{subsec:reasoning_icl}.

% \kun{feel: do we need to break it down by this much?}

\paragraph{ICL gains concentrate on counterfactual examples.}
We next decompose this performance gain of ICL into real and counterfactual examples in the complex setting. As shown in Figure \ref{fig:data_tasks} (C), the two sets of examples differ only in whether the visually grounded answer agrees with or contradicts the learned prior.
If ICL mainly induces format learning, 0-to-16-shot gains should be similar on both subsets.
If ICL helps suppress \prior dominance, gains should concentrate on counterfactual examples.

\begin{figure}[ht]
    \centering
    \begin{minipage}{\linewidth}
        \raggedright
        \textbf{(a)}~\textbf{Aggregate ICL scaling.}\par\smallskip
        \centering
        \includegraphics[width=\linewidth]{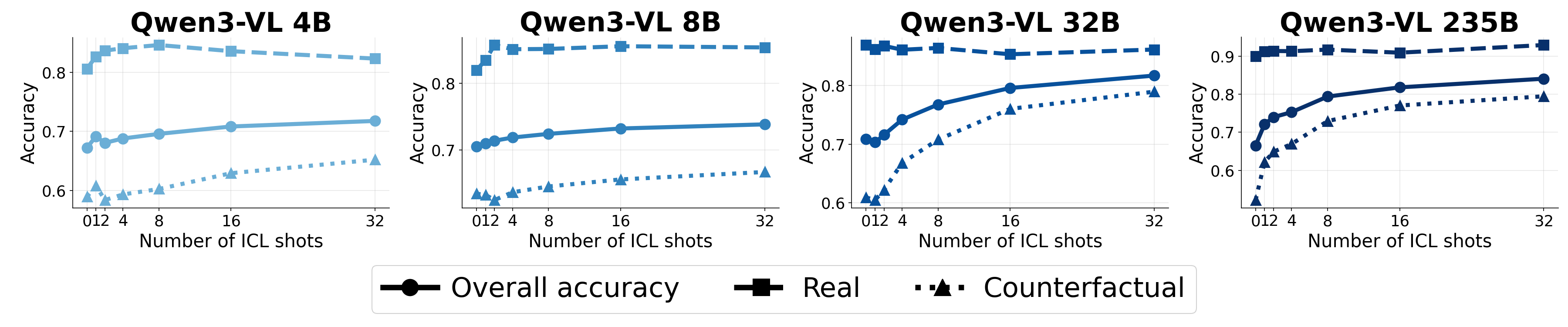}
    \end{minipage}
    \\[1.5ex]
    \begin{minipage}{\linewidth}
        \raggedright
        \textbf{(b)}~\textbf{Per-task ICL diagnostics.}\par\smallskip
        \centering
        \includegraphics[width=\linewidth]{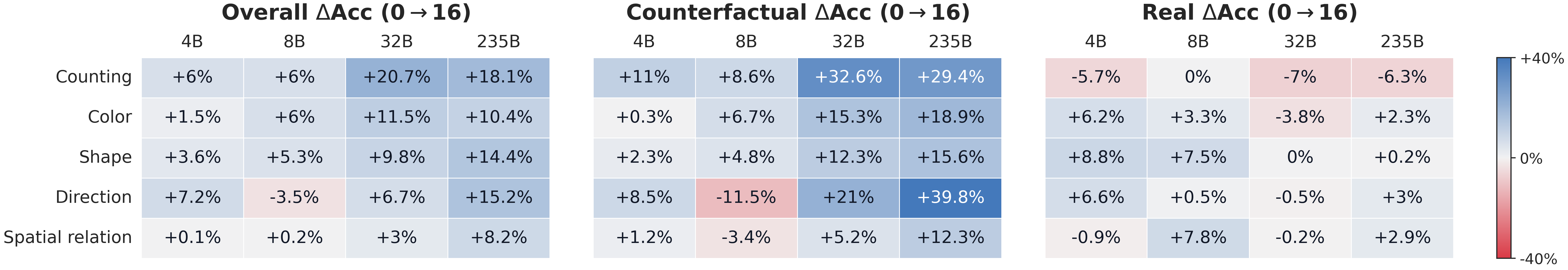}
    \end{minipage}
    \caption{\textbf{Scaling and diagnostic analysis of Qwen3-VL across model sizes on the \emph{complex} \setting.} \textbf{(a)}~Aggregate accuracy on the real and counterfactual subsets with varying numbers of in-context learning (ICL) shots, for the 4B, 8B, 32B, and 235B Qwen3-VL models. \textbf{(b)}~Per-task 0-to-16-shot accuracy change ($\Delta$Acc) on the overall, counterfactual, and real subsets, broken down by task (counting, color, shape, direction, spatial relation) and model size. Positive values (blue) indicate tasks where ICL improves accuracy; negative values (red) indicate degradation.}
    \label{fig:accuracies_cf}
\end{figure}

Figure~\ref{fig:accuracies_cf}(a) supports the \prior-dominance account: the asymmetry is between real and counterfactual gains within each model, not between the 32B and 235B curves.
For the 235B model, 16-shot prompting raises counterfactual accuracy by $+27.2\%$ while leaving real accuracy effectively unchanged ($+2.9\%$).
The 32B model shows the same asymmetry (counterfactual: $+15.1\%$, real: $-1.6\%$).
At 4B and 8B, gains are smaller and less selective, suggesting that smaller models use demonstrations more for format learning than for overriding a learned \prior.
Further results on small models' invariance to shuffled labels suggest that they barely condition on demonstration content~\citep{wei2023largerlanguagemodelsincontext} (Appendix~\ref{app:unlearning}).

Figure~\ref{fig:accuracies_cf}(b) breaks the 0-to-16-shot change down by task. At 32B and 235B, counterfactual gains exceed real gains across all five tasks. 
Counting is the most challenging task in our benchmark and illustrates this most starkly: counterfactual accuracy rises by $+32.6\%$ at 32B and $+29.4\%$ at 235B while real accuracy edges down by $7.0\%$ at 32B and $6.3\%$ at 235B, a pattern we trace in Section~\ref{sec:limitations} to ICL's dependence on a reliable underlying visual capability.
Color identification, shape identification, and spatial relation follow the same pattern while maintaining performance on real examples, with gains scaling monotonically with model capacity.
Object direction is the most scale-dependent task in our benchmark. 
At 32B and 235B it produces the largest absolute counterfactual gain ($+39.8\%$, $+21.0\%$), but at 8B, 16-shot counterfactual accuracy drops below 0-shot.
This scale-dependence is inconsistent with a simple label-availability hypothesis, which would improve all model sizes, and consistent with a capability-dependent override of the \prior.

\paragraph{Gains persist on the expanded benchmark.}
On the expanded complex setting, Qwen3-VL 32B improves from 0 to 16 shots on all five tasks, with gains of 7.2 to 24.8 percentage points (Table~\ref{tab:expanded_benchmark}).
The real share falls from $56.0\%$ to approximately $33\%$, changing the subset weights in overall accuracy; Appendix~\ref{app:expanded_benchmark} reports dataset statistics and the available context standard deviations.

% \xindi{Consider quantifying the gap (e.g. "+X pts on counterfactual vs. +Y pts on real"), this is the paper's strongest empirical finding, own it!:)}
%%%%%%%%%%% subsection 2: mechanistic evaluation %%%%%%%%%%%
\subsection{Mechanistic Evidence for \Prior Override Through Visual Grounding}
\label{sec:mechanistic_evaluation}

Sec.~\ref{sec:performance_evaluation} identifies where visual ICL helps: gains concentrate on counterfactual examples, where the image contradicts the model's learned \prior.
We next test whether demonstrations shift answer generation from image tokens supporting the learned training prior toward tokens supporting the visually correct answer.

% This behavioral result leads to a natural follow-up question: do demonstrations change which visual evidence models use?
% Specifically, we ask whether ICL shifts answer generation away from image tokens that support the learned \prior and toward tokens that ground the correct answer in the current image.

We define \emph{prior cues} as image tokens whose decoded semantics support the answer favored by the learned \prior, and \emph{grounding cues} as tokens whose decoded semantics support the correct answer in the current image. 
To identify these cues, we use logit lens~\citep{nostalgebraist2020logitlens} and LatentLens~\citep{krojer2026latentlensrevealinghighlyinterpretable} to assign approximate semantic labels to image-token representations.
From the running three-legged peacock example in Figure~\ref{fig:1_diagram} (c), tokens decoded as ``peacock'' are prior cues, while tokens decoded as visible legs are grounding cues.
For object direction and spatial relation, no single prior answer is defined, so we use the object named in the caption as a proxy prior cue.
This cue may correlate with the answer without specifying a direct shortcut.
In general, prior and grounding cues need not be mutually exclusive: a single visual token can decode to both kinds of semantics. 
We resolve such cases by assigning any overlapping token to the prior-cue set (Appendix~\ref{app:mechanistic_details}), so the two cue sets are disjoint by construction.

\begin{figure}[t]
    \centering
    \includegraphics[width=0.9\linewidth]{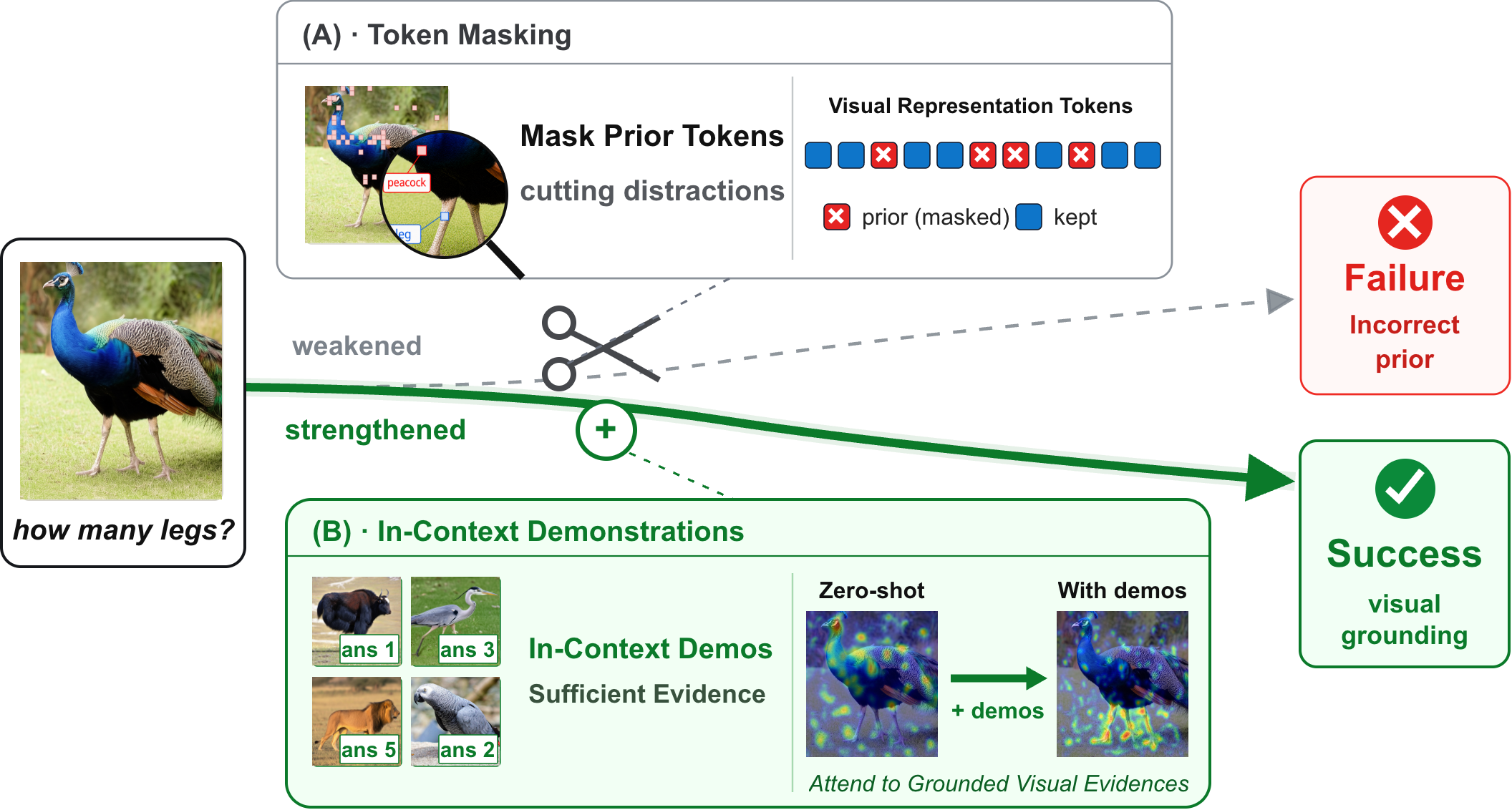}
    \caption{\textbf{Two complementary routes to visual grounding.}
    \textbf{(A)} Masking visual tokens that encode prior cues weakens the learned-\prior pathway.
    \textbf{(B)} In-context demonstrations strengthen attention to image-grounded evidence.
    Both interventions shift the prediction from a prior-driven error toward the visually grounded answer.}
    \label{fig:4_2_diagram}
\end{figure}

With this setup, we test our hypothesis that visual ICL works by reweighting evidence: as the number of demonstrations grows, answer generation should pull less signal from prior-cue tokens and more signal from grounding-cue tokens.
The token-disjoint construction above lets us measure each pull separately.
Figure~\ref{fig:4_2_diagram} lays out the two tests we use: an attention analysis that compares 0-shot and 16-shot weight on the two cue groups, and a masking ablation that asks whether removing prior-cue tokens reproduces the effect of demonstrations.

\paragraph{Demonstrations shift attention toward grounding cues.}
\label{par:attention_shift}
We first test whether demonstrations change the visual evidence used during answer generation.
If ICL overrides \prior dominance by selecting a visually grounded composition, then answer tokens should attend less to prior cues and more to grounding cues.
We measure this shift by comparing answer-token attention under 0-shot and 16-shot prompting, averaging attention over prior cues, grounding cues, and all other image tokens.

\begin{figure}[t!]
    \centering
    \begin{minipage}[t]{0.51\linewidth}
        \vspace{0pt}
        \centering
        {\small\textbf{(a) Demonstrations shift answer-token attention}}
        \vspace{0.1em}

        \includegraphics[width=\linewidth]{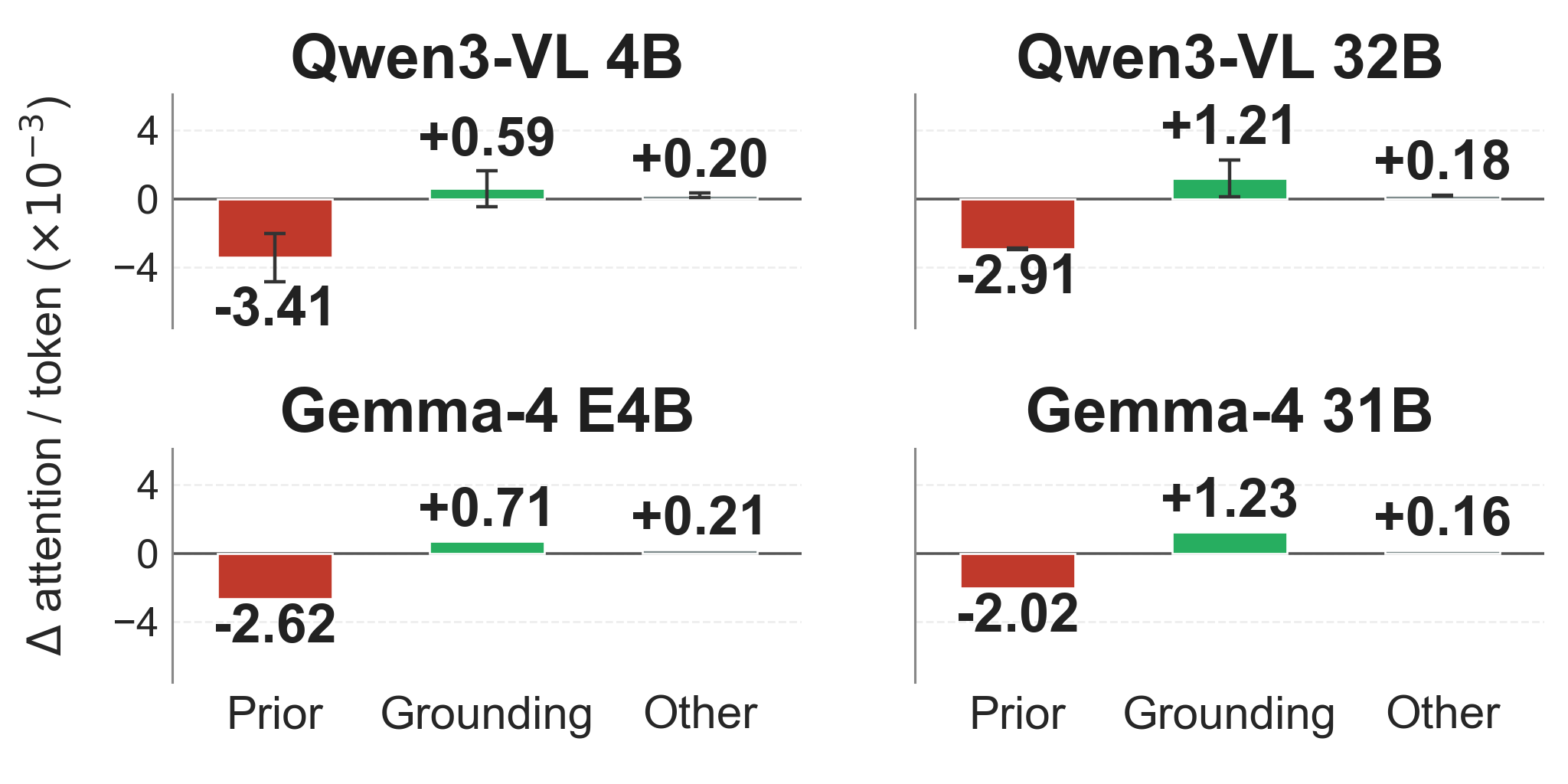}
        \phantomsubcaption
        \label{fig:4_2_attention}
    \end{minipage}\hfill
    \begin{minipage}[t]{0.48\linewidth}
        \vspace{0pt}
        \centering
        {\small\textbf{(b) Masking effects across model families}}
        \vspace{0.1em}

        \footnotesize
        \setlength{\tabcolsep}{2pt}
        \renewcommand{\arraystretch}{1.02}
        \begin{tabular*}{\linewidth}{@{}l@{\extracolsep{\fill}}rrrr@{}}
            \toprule
            & \multicolumn{2}{c}{\textbf{Qwen3-VL 32B}} & \multicolumn{2}{c}{\textbf{Gemma-4 31B}} \\
            \cmidrule(lr){2-3}\cmidrule(l){4-5}
            \textbf{Task} & \textbf{Random} & \textbf{Prior} & \textbf{Random} & \textbf{Prior} \\
            \midrule
            Counting & $-3.9\%$ & $+14.2\%$ & $-4.4\%$ & $+17.9\%$ \\
            Color ID & $+3.5\%$ & $+15.0\%$ & $0.0\%$ & $+9.2\%$ \\
            Shape ID & $-7.2\%$ & $+6.2\%$ & $-0.5\%$ & $+3.2\%$ \\
            Direction & $-3.2\%$ & $+6.1\%$ & $-6.7\%$ & $+2.6\%$ \\
            Spatial rel. & $+0.5\%$ & $+3.4\%$ & $-1.9\%$ & $+1.7\%$ \\
            \bottomrule
        \end{tabular*}
        \vspace{0.2em}

        {\tiny\emph{Accuracy change from each model's 0-shot baseline, in percentage points.}}
        \phantomsubcaption
        \label{tab:4_2_table}
    \end{minipage}
    \caption{\textbf{Demonstrations reduce reliance on prior cues.}
    (a) From 0-shot to 16-shot, VLMs shift answer-token attention away from prior cues and toward grounding cues.
    (b) On the complex setting with greedy decoding, prior-cue masking improves accuracy over both zero-shot and matched random-token masking for Qwen3-VL 32B and Gemma-4 31B on all five tasks.
    Table~\ref{tab:cross_family_mechanistic} reports all attention shifts and full masking accuracies.
    }
    \label{fig:4_2_qualitative}
    % \vspace{-5pt}
\end{figure}

Figure~\ref{fig:4_2_qualitative}(a) shows this redistribution for Qwen3-VL 4B and 32B.
Both models attend less to prior cues and more to grounding cues after demonstrations.
Gemma-4 E4B and 31B show the same pattern (Table~\ref{tab:cross_family_mechanistic}(a)).
These results are consistent with demonstrations redirecting answer generation toward visual evidence already represented in the image tokens.

% Qualitative result for Section 4.2.
% Left: four-model grid of attention shifts from 0-shot to 16-shot.
% Right: zero-shot masking effects for Qwen3-VL 32B and Gemma-4 31B.
% Sources: rebuttal/R2R.md, Table R2.1; rebuttal/R2D.md, Tables 1--2.

\paragraph{Masking prior cues reduces prior-driven errors.}
\label{par:prior_masking}
The attention analysis shows that demonstrations redirect answer generation toward grounding cues, but attention shifts alone do not establish whether prior cues \emph{cause} counterfactual errors.
We therefore run two matched-budget ablations that block visual tokens during answer generation.
The \emph{prior-cue mask} blocks tokens flagged by LatentLens~\citep{krojer2026latentlensrevealinghighlyinterpretable} as decoding to prior-cue semantics; the \emph{random mask} blocks the same number of visual tokens drawn uniformly from non-cue locations.
If zero-shot errors are driven by the semantic content of prior cues rather than by visual-token budget in general, the prior-cue mask should reduce the pull toward the prior-favored answer while the random control should not.

Figure~\ref{fig:4_2_qualitative}(b) compares masking effects for Qwen3-VL 32B and Gemma-4 31B on the \emph{complex} \setting with greedy decoding.
Prior-cue masking improves zero-shot accuracy and outperforms the matched random mask on all five tasks for both models.
For Qwen3-VL 32B, the gains are $14.2$ percentage points on counting, $15.0$ on color identification, and $6.2$ on shape identification.
Gemma-4 31B gains $17.9$, $9.2$, and $3.2$ points, respectively; its counting accuracy rises from $40.9\%$ to $58.8\%$.
Object direction and spatial relation also improve in both families, although the proxy cue makes the evidence less direct for these tasks.
Table~\ref{tab:cross_family_mechanistic}(b) gives the full accuracies, including the corresponding 16-shot results.

The two model families provide consistent evidence for the proposed mechanism.
Demonstrations shift attention from prior cues toward grounding cues, while the masking controls support a contribution of the identified prior cues to zero-shot errors.
Together, these findings support the view that visual ICL shifts inference from $\sprior$ toward $\sground$ by changing how the model uses available visual evidence (Sec.~\ref{sec:formal_definition}).
% This mechanism also predicts a limit: ICL should help when the grounding \composition is available, but not when that \composition is weak or absent.
%%%%%%%%%%

%%%%%%%%%%% subsection 3: limits of ICL %%%%%%%%%%%
\subsection{Visual ICL Relies on Reliable Underlying Visual Capabilities}
\label{sec:limitations}
We have shown that visual ICL helps when VLMs have access to relevant visual evidence but a \prior dominates the answer.
These results raise a natural follow-up question about its limits: does ICL still help when the underlying visual capability itself is unreliable?

% Behaviorally, gains concentrate on counterfactual examples.
% Mechanistically, demonstrations redirect answer generation from prior cues toward grounding cues.

\begin{figure}[t!]
    \centering
    \includegraphics[width=0.9\linewidth]{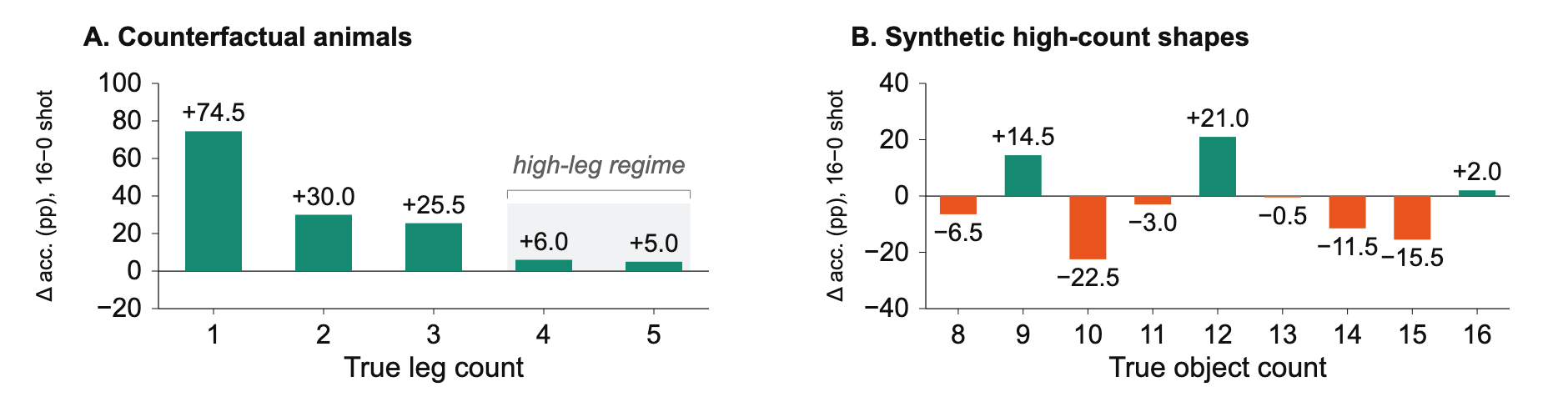}
    \caption{\textbf{ICL gains weaken for harder counting cases.} The plot shows the accuracy change from 0-shot to 16-shot prompting.}
    \label{fig:4_3_diagram}
\end{figure}

\paragraph{ICL gains weaken at higher leg counts.}
We first examine this question within counterfactual counting examples, breaking down ICL gains by the number of animal legs (Figure~\ref{fig:4_3_diagram}(a)).
If prior conflict were the only obstacle, demonstrations should help uniformly across these examples, because each image still contains a misleading animal prior and a visually correct answer.
Instead, ICL helps less for higher leg counts, especially 4- and 5-leg cases, where the model must rely on less robust visual counting.
Overriding the \prior is therefore not sufficient on its own, which motivates a stronger hypothesis: ICL may help VLMs select a reliable capability that is already available, but it does not appear to make an unreliable capability robust given a few demonstrations.

\paragraph{Demonstrations do not make high-cardinality counting robust.}
Counting offers a direct test of this hypothesis because high-cardinality counting remains difficult for VLMs even when objects are simple and clearly visible\citep{bagaria2026recursive,wybitul2024vista,guo2025visionlanguagemodelcantcount,alghisi2025dereconstructingvlmsreasoningcounting}.
We therefore test counting on 450 synthetic 2D-shape images, with 50 examples for each count from 8 through 16 (Table~\ref{tab:dataset_stats}).
These examples remove animal identity, biological plausibility, and learned animal priors, so the main challenge is the counting capability itself.
If demonstrations teach robust high-cardinality counting, then demonstrations for a given count should consistently improve accuracy on test examples with that count.
Contrary to this prediction, ICL gains vary across counts (Figure~\ref{fig:4_3_diagram}(b)), and some counts, such as 10 and 15, fall below zero-shot performance.
Even after removing the confounding effect of learned priors, demonstrations do not reliably improve high-cardinality counting.
These results support a capability-selection view: ICL can redirect a reliable capability suppressed by a prior, but does not consistently strengthen an unreliable capability from a few demonstrations.

\section{Conclusion}
We revisited and challenged the assumption that visual grounding failures in VLMs reflect missing visual capabilities. 
Using a controlled benchmark that separates prior-consistent from prior-conflicting examples and supports label-balanced demonstrations, we show that many of these failures are better understood as \emph{\prior dominance}, where the model encodes the relevant visual evidence and the ability to utilize it but defaults to a stronger learned expectation when answering. 
Task-relevant multimodal demonstrations reliably suppress this effect on prior-conflicting examples, and our mechanistic analyses confirm that demonstrations shift the model's reliance from priors toward visually grounded evidence. 
However, demonstrations do not reliably strengthen weak capabilities. This distinction also matters beyond accuracy. When a VLM defaults to priors instead of perception, those priors can encode stereotypical biases; identifying when a failure is correctable by re-anchoring models to visual evidence is essential for fairness in downstream applications~\citep{seyyed2021underdiagnosis,raj2025vignette,asadi2026mirageillusionvisualunderstanding}.

These findings reframe a subset of visual grounding failures as test-time correctable. 
ICL thereby serves as a diagnostic that separates \emph{\prior dominance}, which test-time intervention can address, from capability absence, which still calls for changes to data, training, or architecture~\citep{gemmateam2025gemma3technicalreport,meng2024deepstack,bai2025qwen3vltechnicalreport}.
Furthermore, visual ICL is a training-free but not compute-free path to adaptation. The interleaved multimodal 32-shot setting consumes substantial context, increases memory use, and adds latency.
Practical continual-adaptation systems will therefore need to select demonstrations carefully, compress or cache visual contexts, or distill successful contexts into model weights~\citep{snell2022learning,hubotter2026reinforcement, shenfeld2026selfdistillationenablescontinuallearning,zhao2026selfdistilledreasoneronpolicyselfdistillation}.

\begin{ack}
  This work is supported by the Princeton First Year Fellowship, the Francis Robbins Upton Fellowship awarded to SC, and the Apple Scholars in AI/ML PhD Fellowship awarded to XW.
  We are grateful to Allison Chen and Tyler Zhu for insightful discussions and valuable feedback that helped shape this work. 
\end{ack}

% References
% \bibliographystyle{unsrt}
\bibliographystyle{unsrtnat}
\bibliography{references}

% Appendix
\appendix
% Appendix
\section{Appendix}

\begingroup
\setcounter{tocdepth}{3}
\etocsettocstyle{\subsection*{Appendix Contents}\small}{}
\etocsetstyle{section}{}{}{}{}
\etocsetstyle{subsection}
  {\par\addvspace{0.35em}}
  {}
  {\noindent{\bfseries\etocifnumbered{\etocnumber\quad}{}\etocname}\nobreak
   \leaders\hbox to 0.8em{\hss.\hss}\hfill\etocpage\par}
  {}
\etocsetstyle{subsubsection}
  {}
  {}
  {\noindent\hspace*{1.5em}\etocifnumbered{\etocnumber\quad}{}\etocname\nobreak
   \leaders\hbox to 0.8em{\hss.\hss}\hfill\etocpage\par}
  {}
\localtableofcontents
\endgroup

% Related Work
\subsection{Related Work}
\label{sec:related_work}

\subsubsection{In-Context Learning in Language Models}

Large language models are strongly shaped by their training distribution.
They are sensitive to pretraining frequencies, token statistics, and output-format priors, even when a prompt asks for a different behavior~\citep{mccoy2023embersautoregressionunderstandinglarge,razeghi2022impactpretrainingtermfrequencies,liu2026formatpriorquantifyinganalyzing}.
In-context learning offers one way to change this behavior at inference time.
Since the first few-shot results in large language models~\citep{brown2020languagemodelsfewshotlearners,wei2022emergentabilitieslargelanguage}, several studies have described ICL as a form of implicit task inference, where demonstrations provide evidence about the task and shift the model toward predictors that explain the examples~\citep{xie2021explanation,min2022rethinking,wakayama2025incontextlearningprovablybayesian}.
This shift can be strong enough to override default behavior.
Language models can learn counterintuitive label mappings from context~\citep{wei2023largerlanguagemodelsincontext}, suppress unwanted behavior~\citep{pawelczyk2024incontextunlearninglanguagemodels}, and infer task rules from demonstrations~\citep{pan2023incontextlearninglearnsincontext,kossen2024incontextlearninglearnslabel,lampinen2025generalizationlanguagemodelsincontext}.

Because ICL leaves model weights fixed and changes only the context, it serves as a probe of what a model already knows: if demonstrations shift behavior, the underlying capability was present but not selected by default.
We adopt this diagnostic framing for vision-language models, where a shift from a prior-favored answer to a visually grounded answer would indicate that the visual capability was already available.

\subsubsection{Visual Failures in Vision-Language Models: Capability or Prior?}

Vision-language models still fail on tasks that require precise visual understanding, including chart reasoning~\citep{zhang2024mathversedoesmultimodalllm}, optical illusions~\citep{shahgir2024illusionvqachallengingopticalillusion}, inductive visual reasoning~\citep{chollet2026arcagi2newchallengefrontier}, and basic visual perception~\citep{chen2026babyvisionvisualreasoninglanguage}.
Most existing remedies treat these failures as missing visual capability and respond by editing attention or internal representations~\citep{kang2025yourground,zhao2025tellground,wan2024contrastiveground,li2026makecama}, by altering the training recipe or visual encoder~\citep{qwen2025qwen25technicalreport,bai2025qwen3vltechnicalreport,meng2024deepstack,gemmateam2025gemma3technicalreport}, or by preparing the input through cropping, zooming, or visual prompting~\citep{wu2024v,shen2025zoomeye,shtedritski2023doesredcircle}.
These approaches differ in mechanism but share an assumption: the model cannot yet use the relevant visual evidence.

A separate line of work documents that VLMs often rely on \priors instead of the image.
Some models answer visual questions from linguistic associations even when the image gives contrary evidence~\citep{wu2023rolelinguisticpriorsmeasuring,asadi2026mirageillusionvisualunderstanding}, and predictions are highly sensitive to textual cues that pull the answer away from visual evidence~\citep{luo2025probingvisuallanguagepriors,lee2025vlindbenchmeasuringlanguagepriors}.
The bias extends beyond language: VLMs have clear preferences on which visual information to incorporate~\citep{allen2025incontextinductivebiasesvisionlanguage} and exploit shortcut \compositions when solving visual tasks~\citep{vo2025visionlanguagemodelsbiased,irawan2025visionlanguagemodelsconfused, golovanevsky-etal-2025-pixels-priors}.
Recent analyses further show that useful visual information can be present in the model yet not used~\citep{fu2025hiddenplainsightvlms,liu2025visualrepresentationsinsidelanguage}.

Our work is motivated by a question raised by \citet{luo2025probingvisuallanguagepriors, irawan2025visionlanguagemodelsconfused, golovanevsky-etal-2025-pixels-priors}: can VLMs overcome \priors when supplied with in-context demonstrations that anchor their predictions in visual evidence? While \citet{vo2025visionlanguagemodelsbiased} report that demonstrations yield little benefit, we argue that these null results stem from inadequate experimental controls. In particular, the insufficient number of demonstrations and limited diversity of answer labels hinder a model's ability to generalize.

\subsubsection{Multimodal In-Context Learning}

Models that interleave text and images make multimodal in-context learning possible~\citep{alayrac2022flamingovisuallanguagemodel}.
Whether current VLMs perform genuine visual ICL remains contested.
Several studies report that multimodal ICL is driven mainly by text, prompt format, or context cues rather than by demonstrated visual evidence~\citep{baldassini2024makesmultimodalincontextlearning,chen2024multimodallargelanguagemodels,li2026makecama}, and that apparent gains can come from shallow strategies such as copying, matching, or majority voting~\citep{baldassini2024makesmultimodalincontextlearning,huang2025mimickingreasoningrethinkingmultimodal}.
Multimodal ICL also lags text-only ICL on average~\citep{wang2026multimodalincontextlearninglags} and is brittle under changes in perception, reasoning, or context length~\citep{zong2025vliclbenchdevildetails,chen2025truemultimodalincontextlearning,li2025advancingmultimodalincontextlearning}.

The most relevant work to our study demonstrates that VLMs possess some capacity to overcome semantic priors and follow unseen tasks from in-context demonstrations~\citep{xu-etal-2025-introspection-analysis-icl}. However, we argue that the ICL settings in this study inadvertently invite shortcut learning rather than true visual grounding. Simple strategies, such as flipping answer labels or adopting an ``I don't know'' response format, constitute low-effort shortcuts that models can acquire directly from demonstrations without any reliance on visual evidence. Beyond this, the study's scope is largely limited to demonstration retrieval in general VQA benchmarks, with minimal control over task-specific factors central to our investigation.

\subsection{Further Dataset and Evaluation Settings}

\subsubsection{Dataset}

\paragraph{Task Definitions and Setup.}
This appendix expands the task and \setting definitions from Section~\ref{sec:dataset_tasks}.
We use the main-paper terminology throughout: each example has a \emph{grounded answer} determined by the image and, when applicable, a \emph{prior answer} favored when the model ignores that evidence.
In the \emph{complex} \setting, \emph{real} examples are prior-consistent because these answers agree, while \emph{counterfactual} examples are prior-conflicting because they disagree.
The five closed-label tasks are counting, color identification, shape identification, object direction, and spatial relation.
The \emph{simple} \setting tests the visual operation on abstract stimuli, the \emph{no-prior} \setting replaces target names with placeholders so demonstrations specify the task rule, and the \emph{complex} \setting tests \prior override.
Together, the tasks cover two shortcut regimes relevant to \prior dominance: counting, color identification, and shape identification test language or world-knowledge priors tied to object identity, while object direction and spatial relation test visual-configuration priors tied to common poses and layouts.
% Representative stimuli are shown in Figure~\ref{fig:benchmark_full}.

\subparagraph{Counting.}
Following~\citet{vo2025visionlanguagemodelsbiased}, this task asks for the number of visible animal legs in the \emph{complex} \setting.
Real examples match common species leg counts, while counterfactual examples change the visible count.
The \emph{simple} and \emph{no-prior} \settings use synthetic circle-counting images; \emph{no-prior} replaces the named target with the placeholder ``target shapes.''
Answer labels are integers in $[1,8]$ for the \emph{simple} and \emph{no-prior} \settings and $[1,5]$ for the \emph{complex} \setting.

\subparagraph{Color Identification.}
Following~\citet{luo2025probingvisuallanguagepriors}, this task asks for the displayed color of the target object.
The \emph{complex} \setting uses animals whose visible color either matches the species prior or conflicts with it.
The \emph{simple} and \emph{no-prior} \settings use synthetic colored shapes; \emph{no-prior} fixes the target as the single circle and asks for the color of the ``target shape.''
All three \settings use \{green, orange, brown, black, pink, red\}, derived from the canonical colors of the six target animals in the \emph{complex} \setting.

\subparagraph{Shape Identification.}
Following~\citet{luo2025probingvisuallanguagepriors}, this task asks for the displayed shape of the target object.
The \emph{complex} \setting uses traffic signs whose rendered shape either matches the sign prior or conflicts with it.
The \emph{simple} and \emph{no-prior} \settings use synthetic 2D shapes; \emph{no-prior} fixes the hidden target as the red object and asks for the shape of the ``target object.''
Answer labels are \{circle, triangle, rectangle, pentagon, octagon, diamond\}.

\subparagraph{Object Direction.}
Following~\citet{jung2025rightrightenhancingobject}, this task asks for the camera-relative orientation of a target object.
The \emph{complex} \setting stresses visual-configuration priors by adding uncommon up and down orientations to the front, back, left, and right directions covered by the source benchmark.
The \emph{simple} and \emph{no-prior} \settings use SVG objects with red and green markers indicating the front-facing direction; \emph{no-prior} uses more abstract non-arrow targets.
Answer labels are \{front, back, left, right, up, down\}.

\subparagraph{Spatial Relation.}
Following~\citet{kamath2023whatsupvisionlanguagemodels}, this task asks for the relation of an ordered target object to a reference object.
The \emph{complex} \setting preserves the object pair while changing whether the depicted relation agrees with a frequent spatial layout.
The \emph{simple} and \emph{no-prior} \settings use synthetic colored shapes; \emph{no-prior} fixes object~A as a red circle and object~B as a blue triangle, then queries the relation using only the placeholders object~A and object~B.
All \settings use the same answer labels: \{left, right, front, behind, top, bottom\}.

\begin{figure}[t]
\centering
\begin{tcolorbox}[
    enhanced,
    colback=gray!3,
    colframe=gray!60,
    boxrule=0.6pt,
    arc=2pt,
    left=8pt, right=8pt, top=6pt, bottom=6pt,
    title=\textbf{Prompts for Tasks},
    fonttitle=\bfseries,
    coltitle=white,
    colbacktitle=gray!70,
]
\ttfamily\small
\raggedright

\textbf{\textsf{Counting}}\par\vspace{2pt}
\textcolor{blue!60!black}{\textbf{[Complex]}}~~How many legs does this animal have? Answer with a number in curly brackets. e.g., \{4\}.\par\vspace{2pt}
\textcolor{green!55!black}{\textbf{[Simple]}}~~How many circles are there? Answer with a number in curly brackets. e.g., \{6\}.\par\vspace{2pt}
\textcolor{orange!70!black}{\textbf{[No-prior]}}~~How many target shapes are there? Answer with a number in curly brackets. e.g., \{6\}.

\vspace{4pt}{\color{gray!40}\hrule}\vspace{4pt}

\textbf{\textsf{Color Identification}}\par\vspace{2pt}
\textcolor{blue!60!black}{\textbf{[Complex]}}~~What color is the animal in this image? Choose from: \{green, orange, brown, black, pink, red\}. Answer with the color in curly brackets, e.g., \{red\}.\par\vspace{2pt}
\textcolor{green!55!black}{\textbf{[Simple]}}~~What color is the \textlangle shape\textrangle? Choose from: \{green, orange, brown, black, pink, red\}. Answer with the color in curly brackets, e.g., \{red\}.\par\vspace{2pt}
\textcolor{orange!70!black}{\textbf{[No-prior]}}~~What color is the target shape? Choose from: \{green, orange, brown, black, pink, red\}. Answer with the color in curly brackets, e.g., \{red\}.

\vspace{4pt}{\color{gray!40}\hrule}\vspace{4pt}

\textbf{\textsf{Shape Identification}}\par\vspace{2pt}
\textcolor{blue!60!black}{\textbf{[Complex]}}~~What shape is the traffic sign in this image? Answer with one of: \{circle, triangle, rectangle, pentagon, octagon, diamond\}.\par\vspace{2pt}
\textcolor{green!55!black}{\textbf{[Simple]}}~~What shape is the \textlangle color\textrangle~object? Choose from: \{circle, triangle, rectangle, pentagon, octagon, diamond\}. Answer with the shape name in curly brackets, e.g., \{circle\}.\par\vspace{2pt}
\textcolor{orange!70!black}{\textbf{[No-prior]}}~~What shape is the target object? Choose from: \{circle, triangle, rectangle, pentagon, octagon, diamond\}. Answer with the shape name in curly brackets, e.g., \{circle\}.

\vspace{4pt}{\color{gray!40}\hrule}\vspace{4pt}

\textbf{\textsf{Object Direction}}\par\vspace{2pt}
\textcolor{blue!60!black}{\textbf{[Complex]}}~~From the perspective of the camera, which orientation is the object in the photo facing? Choose from: \{front\}, \{back\}, \{left\}, \{right\}, \{up\}, \{down\}. Answer with the direction word in curly brackets, e.g., \{front\}.\par\vspace{2pt}
\textcolor{green!55!black}{\textbf{[Simple]}}~/~\textcolor{orange!70!black}{\textbf{[No-prior]}}~~From the perspective of the camera, which orientation is the target object in the photo facing? Choose from: \{front\}, \{back\}, \{left\}, \{right\}, \{up\}, \{down\}. Answer with the direction word in curly brackets, e.g., \{front\}.

\vspace{4pt}{\color{gray!40}\hrule}\vspace{4pt}

\textbf{\textsf{Spatial Relation}}\par\vspace{2pt}
\textcolor{blue!60!black}{\textbf{[Complex]}}~~What is the spatial relation of the \textlangle objA\textrangle~relative to the \textlangle objB\textrangle? Choose from the following options: \{left\}, \{right\}, \{front\}, \{behind\}, \{top\}, \{bottom\}. Answer with only the relation in curly brackets, e.g., \{left\}.\par\vspace{2pt}
\textcolor{green!55!black}{\textbf{[Simple]}}~~What is the spatial relation of the \textlangle colorA~shapeA\textrangle~relative to the \textlangle colorB~shapeB\textrangle? Choose from the following options: \{left\}, \{right\}, \{front\}, \{behind\}, \{top\}, \{bottom\}. Answer with the relation in curly brackets, e.g., \{left\}.\par\vspace{2pt}
\textcolor{orange!70!black}{\textbf{[No-prior]}}~~What is the spatial relation of object A relative to object B? Choose from the following options: \{left\}, \{right\}, \{front\}, \{behind\}, \{top\}, \{bottom\}. Answer with the relation in curly brackets, e.g., \{left\}.
\end{tcolorbox}
\caption{Prompts used for querying models across tasks and \settings.}
\label{fig:benchmark_prompts}
\end{figure}

\paragraph{Dataset Statistics.}
\label{app:dataset_stats}

Table~\ref{tab:dataset_stats} reports sources, labels, and counts for all tasks, with real/counterfactual statistics for each complex task.
It also lists the counterfactual counting subset and the 450 high-count shape images used in Section~\ref{sec:limitations}.

% Original complex counts: rebuttal/R1R.md, Table R1.3.
% High-count shape statistics: rebuttal/R1R.md, Section 5.
\begin{table}[!htbp]
\centering
\caption{Dataset statistics. (a) Original benchmark sources, labels, and counts, including real and counterfactual (CF) subsets. The real/CF split applies only to complex examples. Citations identify image sources or, for color and shape, task designs. Procedural images use Pillow or SVG; generated, regenerated, and edited images use Gemini. Regeneration reproduces a source image; HV denotes human validation. (b) Counting datasets used in Section~\ref{sec:limitations}; high-count shapes have 50 images per label.}
\label{tab:dataset_stats}
\small
\setlength{\tabcolsep}{3pt}
\textbf{(a) Sources, Labels, and Original Counts}\par\smallskip
\begin{tabular}{@{}llp{3.2cm}p{1.8cm}ccccc@{}}
\toprule
\textbf{Task} & \textbf{\Setting} & \textbf{Source} & \textbf{Labels} & \textbf{Count} & \textbf{Real} & \textbf{CF} & \textbf{Edited} & \textbf{HV} \\
\midrule
Counting & Simple & Procedural (Pillow) & $\{1, \dots, 8\}$ & 320 & & & & \\
Counting & No-prior & Procedural (Pillow) & $\{1, \dots, 8\}$ & 320 & & & & \\
Counting & Complex & VLMs are Biased (regenerated)~\citep{vo2025visionlanguagemodelsbiased} & $\{1, \dots, 5\}$ & 200 & 90 & 110 & \checkmark & \checkmark \\
\midrule
Color ID & Simple & Procedural (Pillow) & \{G, O, Br, Bk, Pk, R\} & 240 & & & & \\
Color ID & No-prior & Procedural (Pillow) & \{G, O, Br, Bk, Pk, R\} & 240 & & & & \\
Color ID & Complex & Generated + Edited; task design~\citep{luo2025probingvisuallanguagepriors} & \{G, O, Br, Bk, Pk, R\} & 240 & 120 & 120 & \checkmark & \checkmark \\
\midrule
Shape ID & Simple & Procedural (Pillow) & 6 shapes & 240 & & & & \\
Shape ID & No-prior & Procedural (Pillow) & 6 shapes & 240 & & & & \\
Shape ID & Complex & Generated + Edited; task design~\citep{luo2025probingvisuallanguagepriors} & 6 shapes & 240 & 120 & 120 & \checkmark & \checkmark \\
\midrule
Direction & Simple & Procedural (SVG) & 6 directions & 240 & & & & \\
Direction & No-prior & Procedural (SVG) & 6 directions & 240 & & & & \\
Direction & Complex & EgoOrientBench (regenerated)~\citep{jung2025rightrightenhancingobject} & 6 directions & 240 & 160 & 80 & \checkmark & \checkmark \\
\midrule
Spatial & Simple & Procedural (Pillow) & 6 relations & 240 & & & & \\
Spatial & No-prior & Procedural (Pillow) & 6 relations & 240 & & & & \\
Spatial & Complex & WhatsUp; COCO-/GQA-spatial~\citep{kamath2023whatsupvisionlanguagemodels,lin2015microsoftcococommonobjects} (regenerated) & 6 relations & 240 & 160 & 80 & \checkmark & \checkmark \\
\bottomrule
\end{tabular}
\par\medskip
\textbf{(b) Further Counting Studies (Section~\ref{sec:limitations})}\par\smallskip
\begin{tabular}{@{}llcrl@{}}
  \toprule
  \textbf{Dataset} & \textbf{Source} & \textbf{Labels} & \textbf{Count} & \textbf{Study} \\
  \midrule
  Animal-leg counting & CF counting subset above & $\{1,\dots,5\}$ & 110 & Figure~\ref{fig:4_3_diagram}(a) \\
  High-count shapes & Synthetic 2D shapes & $\{8,\dots,16\}$ & 450 & Figure~\ref{fig:4_3_diagram}(b) \\
  \bottomrule
\end{tabular}
\end{table}

\paragraph{Training Priors for Object Direction and Spatial Relation.}
\label{app:configuration_priors}

These tasks probe preferences for common poses and layouts, with the preferred answer depending on the object or object pair.
The selected EgoOrientBench entries~\citep{jung2025rightrightenhancingobject} contain only front, back, left, and right labels; up and down each have zero examples.
In the original complex direction split, edited up/down examples form the entire counterfactual subset: $80/240 = 33.3\%$ of the queries.
For spatial relation, the selected entries from each source (WhatsUp VLM, COCO-spatial, and GQA-spatial) also contain zero top and zero bottom examples.
WhatsUp's controlled images use unusual object combinations to reduce object-dependent biases~\citep{kamath2023whatsupvisionlanguagemodels}, but the selected entries still leave these labels uncovered.
These zero counts describe source coverage before counterfactual editing; the constructed benchmark covers all six labels for each task.
In the original complex setting, both tasks contain 160 real and 80 counterfactual examples (Table~\ref{tab:dataset_stats}).

We hypothesize that VLM training similarly favors common configurations, but model-specific training frequencies are unavailable and cannot be inferred from these source counts.
To test this preference, we probe Qwen3-VL 32B directly.
For object direction, we replace the image with object-identity text from its source caption, such as ``A girl carrying an umbrella.''
Table~\ref{tab:configuration_priors} reports zero-shot accuracy with images and with this caption-only input.
With images, the real-to-counterfactual accuracy gap is 47.4 percentage points for direction and 42.0 points for spatial relation.
Caption-only direction accuracy remains above uniform six-label chance ($16.7\%$) on real examples ($42.1\%$), but falls below it on counterfactuals ($2.7\%$).
This asymmetry is consistent with a preference for common configurations without visual evidence.

Both the \emph{simple} and \emph{no-prior} \settings retain each task's six-label answer set.
In the \emph{simple} \setting, Qwen3-VL 32B reaches zero-shot accuracies of $78.2\%$ for direction and $80.8\%$ for spatial relation (Table~\ref{tab:app_cross_family_accuracies}).
These controls argue against a general inability to use the answer sets.
The direction split remains a rare-orientation stress test because its real and counterfactual subsets use disjoint labels.
Their accuracy gap therefore cannot separate orientation rarity from object-dependent priors.

% Source: rebuttal/R1R.md, Tables R1.1 and R1.2.
% Accuracies are percentages; overall values weight 160 real and 80 counterfactual examples.
\begin{table}[!htbp]
  \centering
  \small
  \captionsetup{position=bottom,skip=6pt}
  \setlength{\tabcolsep}{8pt}
  \begin{tabular}{@{}llrrr@{}}
    \toprule
    \textbf{Task} & \textbf{Input} & \textbf{Real} & \textbf{Counterfactual} & \textbf{Overall} \\
    \midrule
    Object Direction & Image & 93.3 & 45.9 & 77.5 \\
                     & Caption Only & 42.1 & 2.7 & 29.0 \\
    \midrule
    Spatial Relation & Image & 83.2 & 41.2 & 69.2 \\
    \bottomrule
  \end{tabular}
  \caption{\textbf{Configuration-prior diagnostics.} Zero-shot accuracy (\%) for Qwen3-VL 32B on the original complex setting. Caption-only input replaces the direction image with object-identity text from its source caption. Overall accuracy weights the 160 real and 80 counterfactual examples in each task.}
  \label{tab:configuration_priors}
\end{table}

\paragraph{Data Construction and Validation.}
\label{app:data_construction}

% Figure showing dataset construction pipeline
\begin{figure}[t]
    \centering
    \includegraphics[width=\linewidth]{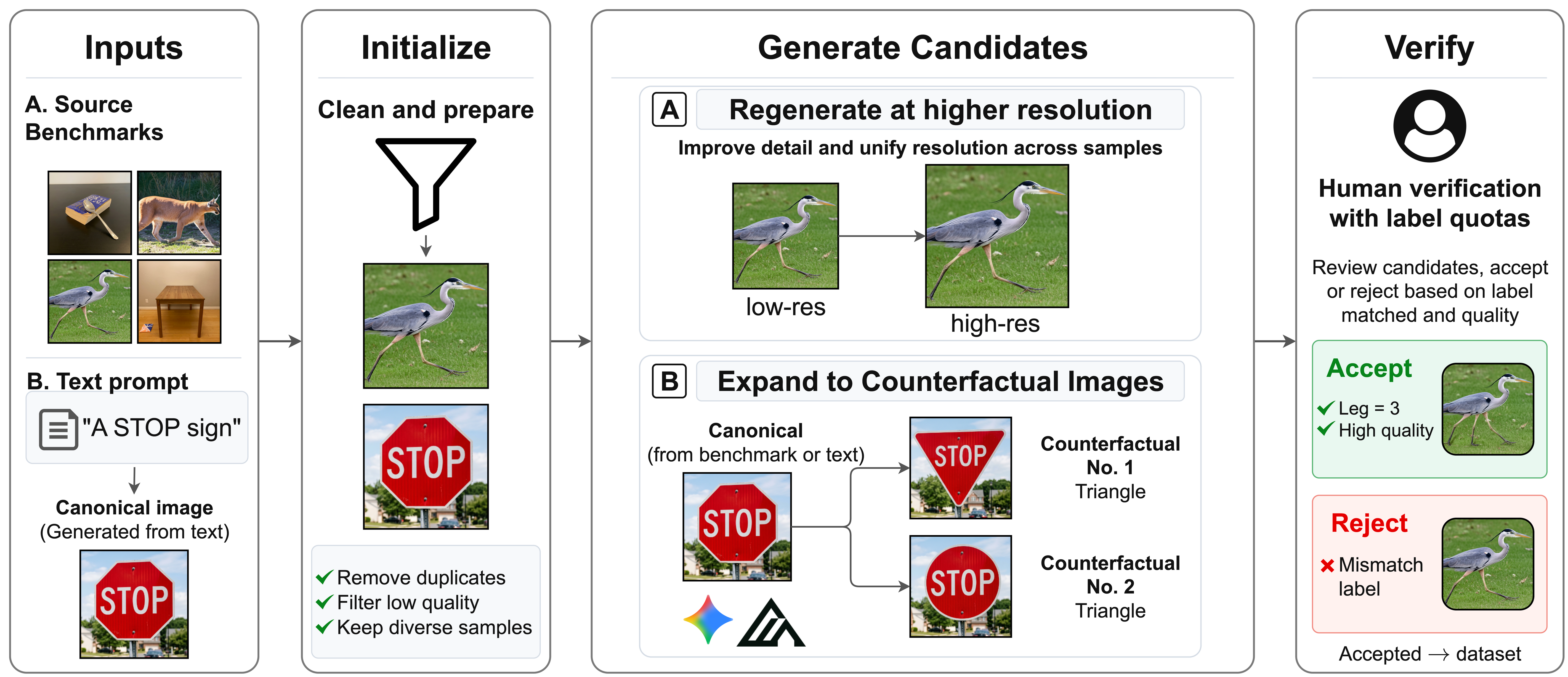}
    \caption{\textbf{Dataset construction.} For each task, images are rendered, generated, regenerated, or edited across the simple, no-prior, and complex settings, then paired with closed answer labels and validation controls for label-balanced ICL evaluation.}
    \label{fig:benchmark_data_construction}
\end{figure}

Our benchmark follows a common construction pipeline across tasks: initialize a candidate pool, expand the pool when prior-conflicting labels are needed, manually verify each candidate, and retain only verified samples in the cleaned benchmark. The pipeline varies by task and \setting (Figure~\ref{fig:benchmark_data_construction}). Synthetic \emph{simple} and \emph{no-prior} \settings are rendered with deterministic programs. The \emph{complex} \settings start from existing benchmark images or generated prior-consistent images, then use targeted image editing to produce prior-conflicting examples. After construction and validation, all retained images are downsampled to $512 \times 512$ pixels.

\subparagraph{\emph{Simple} and \emph{no-prior} \settings.}
All \emph{simple} and \emph{no-prior} images are synthetically rendered with Pillow or SVG on a $512 \times 512$ canvas. No image generation model is used.

\begin{itemize}
    \item \textbf{Counting.} Synthetic scenes contain same-color target circles and non-circle distractors. The \emph{no-prior} prompt replaces ``circles'' with ``target shapes.''
    \item \textbf{Color identification.} Synthetic scenes contain a colored target shape and distractor shapes. The \emph{no-prior} prompt fixes the target as the single circle and asks for the ``target shape.''
    \item \textbf{Shape identification.} Synthetic scenes contain a target shape and distractors in different colors. The \emph{no-prior} prompt fixes the hidden target as the red object and asks for the ``target object.''
    \item \textbf{Object direction.} SVG target objects are rendered with distractors and front-direction markers. The \emph{simple} \setting uses arrow-like targets, and the \emph{no-prior} \setting uses abstract non-arrow targets.
    \item \textbf{Spatial relation.} Two colored shapes are placed in one of six spatial relations, with optional distractors. The \emph{no-prior} prompt fixes object~A as a red circle and object~B as a blue triangle, then refers only to object~A and object~B.
\end{itemize}

\subparagraph{Interpreting the \emph{no-prior} zero-shot baseline.}
The \emph{no-prior} \setting is a task-rule inference control, not the main evidence for \prior override.
Its prompts replace target names with placeholders, so demonstrations provide the hidden binding; this can make zero-shot queries underspecified when distractors are present.
The primary \prior override evidence comes from the \emph{complex} real/counterfactual split, where the queried target is explicit and the grounded answer either agrees with or conflicts with the prior answer.

\subparagraph{\emph{Complex} \settings.}
Images in the \emph{complex} \settings are either prior-consistent real examples or prior-conflicting counterfactual examples.
Their base images come from existing benchmarks or are generated from scratch, and are then regenerated or counterfactually edited to produce the desired real/counterfactual split.
We control for source-domain artifacts by generating or regenerating the prior-consistent images with Gemini rather than comparing raw originals against Gemini-edited counterfactuals.
Thus, the real and counterfactual splits both come from the Gemini construction pipeline and share the same human validation, label balancing, and downsampling steps.
All generated, regenerated, and edited images use Gemini~\citep{gemini_team_google_2023,googledeepmind_gemini31pro_modelcard_2026} at $1024 \times 1024$ resolution unless otherwise noted.

\begin{itemize}
    \item \textbf{Counting.} Base images are drawn from the Animals topic of \citet{vo2025visionlanguagemodelsbiased}, filtered for counting questions with resolution $\geq$1024\,px. Selected base images are regenerated at $1024 \times 1024$ to place prior-consistent examples in the same Gemini output domain as edited examples. For each target leg count in $\{1, 2, 3, 4, 5\}$, we edit base images so the animal has exactly the target number of legs. Edited images are manually verified with a per-label cap of 40, and unverified samples are removed.
    \item \textbf{Color identification.} We generate 10 canonical images per animal for six species with distinct canonical colors (frog/green, tiger/orange, capybara/brown, raven/black, flamingo/pink, scarlet ibis/red) at $512 \times 512$. These canonical Gemini generations form the real split. Each canonical image is then edited into the five non-original colors, producing counterfactual variants where the animal color conflicts with the species prior. Edited images are manually verified with a per-color cap of 40.
    \item \textbf{Shape identification.} We generate 10 canonical images per traffic sign for six sign types with distinct canonical shapes (stop/octagon, yield/triangle, railroad crossing ahead/circle, school zone crossing/pentagon, speed limit/rectangle, warning/diamond) at $512 \times 512$. These canonical Gemini generations form the real split. Each canonical image is then edited into the five non-original shapes. Edited images are manually verified with a per-shape cap of 40.
    \item \textbf{Object direction.} Base images are drawn from EgoOrientBench~\citep{jung2025rightrightenhancingobject}, filtered for \texttt{type=choose} entries with cardinal directions (front, back, left, right). Selected base images are regenerated at $1024 \times 1024$, and these regenerated images form the real split for the original directions. For each of the two missing directions (up, down), we edit base images to reorient the object accordingly (100 per direction). Edited images are manually verified with a per-direction cap of 40 across all six directions.
    \item \textbf{Spatial relation.} Base images and captions are drawn from WhatsUp VLM~\citep{kamath2023whatsupvisionlanguagemodels} (subsets~A and~B), COCO spatial~\citep{lin2015microsoftcococommonobjects}, and GQA spatial. Object pairs and spatial relations are parsed from caption text. The prior-consistent base images are regenerated with Gemini and form the real split. For each base sample, a local language model generates up to two counterfactual captions that change only the spatial relation between the same objects, choosing relations that make the resulting configuration rare or unnatural. Each counterfactual caption is then used to edit the regenerated base image so the objects match the new spatial arrangement. Edited images are manually verified with a per-relation cap of 40 across all six relations.
\end{itemize}

\subparagraph{Human Validation and Filtering.}
\label{app:human_validation}
Two in-house annotators independently label every image in the \emph{complex} \settings, both blind to the intended label.
Each annotator sees the image, task question, and answer options, then selects an answer in the same multiple-choice format used for model evaluation.
We retain an image only if both annotators select the intended label and confirm that it meets the quality criteria.
Any disagreement between annotators or with the intended label leads to rejection rather than adjudication.

Each annotator checks that the target object or relation is visible and unambiguous, distractors leave only one valid answer, and edits preserve object identity and scene realism.
Text, markings, and object parts needed to answer the question must remain legible.
We reject images with ambiguous, occluded, or cropped targets, visible editing boundaries, texture discontinuities, label-changing artifacts, prompt--image mismatches, or blur that prevents reliable judgment.
The task-specific checks are:
\begin{itemize}
  \item \textbf{Counting.} The animal's visible legs must support an unambiguous count. We reject images with unclear leg boundaries or no clear indication of individual legs.
  \item \textbf{Spatial Relation and Object Direction.} The depicted relation or orientation must correspond to a single answer option. We reject mixed directions, such as a person in three-quarter view looking at the camera.
  \item \textbf{Color and Shape Identification.} The main object's displayed color or shape must be unambiguous. We reject generated images that incorrectly mix attributes.
\end{itemize}
After filtering, we cap the retained images per label to improve label balance.
Applying this protocol to the original dataset did not change the retained images.

Table~\ref{tab:human_validation} reports validation statistics for the original counterfactual portion of the \emph{complex} \setting.
Across the five tasks, we accepted 510 of 936 candidates ($54.5\%$) and rejected 426.
These acceptance rates combine label agreement with the image-quality checks above; they do not measure inter-annotator agreement across all candidates.

% Source: rebuttal/R2R.md, Section 1, Table R2.0.
\begin{table}[!htbp]
  \centering
  \caption{Human validation statistics for the original counterfactual portion of the complex setting. Acceptance requires both annotators to select the intended label independently and confirm that the image meets the quality criteria.}
  \label{tab:human_validation}
  \small
  \setlength{\tabcolsep}{8pt}
  \begin{tabular}{@{}lrrrr@{}}
    \toprule
    \textbf{Task} & \textbf{Accepted} & \textbf{Rejected} & \textbf{Candidates} & \textbf{Acceptance (\%)} \\
    \midrule
    Counting         & 110 & 192 & 302 & 36.4 \\
    Color ID         & 120 &  15 & 135 & 88.9 \\
    Shape ID         & 120 &  13 & 133 & 90.2 \\
    Object Direction &  80 &  78 & 158 & 50.6 \\
    Spatial Relation &  80 & 128 & 208 & 38.5 \\
    \midrule
    \textbf{All Tasks} & \textbf{510} & \textbf{426} & \textbf{936} & \textbf{54.5} \\
    \bottomrule
  \end{tabular}
\end{table}

\subparagraph{Counterfactual Generation Failures.}
Acceptance rates are lowest for counting ($36.4\%$), spatial relation ($38.5\%$), and object direction ($50.6\%$), compared with color ($88.9\%$) and shape identification ($90.2\%$).
These rates show the difficulty of generating unambiguous counterfactual images.
They are consistent with generation models favoring familiar attributes and configurations, although the rejection counts also include ambiguity and image-quality failures.

\subparagraph{Generation and editing prompts.}
\label{app:generation_prompts}
Table~\ref{tab:generation_prompts} lists the exact prompts used for image generation and counterfactual editing in \emph{complex} \settings. All prompts are passed to Gemini~\citep{gemini_team_google_2023,googledeepmind_gemini31pro_modelcard_2026}. Placeholders in braces (e.g., \texttt{\{animal\_name\}}) are filled at generation time.
For regeneration from existing images, the source image is supplied as visual input and the instruction asks Gemini to reproduce the scene at the target resolution while preserving object identity, the answer-defining attribute or relation, background, lighting, and camera angle.
This regeneration step leaves the real label unchanged; it only places prior-consistent examples in the same image-model output domain as counterfactual examples.

\begin{table}[ht]
\centering
\caption{Generation and editing prompts used for image construction in the \emph{complex} \setting.}
\label{tab:generation_prompts}
\small
\begin{tabular}{p{2.2cm}p{1.5cm}p{\dimexpr\linewidth-3.7cm-6\tabcolsep\relax}}
\toprule
\textbf{Task} & \textbf{Stage} & \textbf{Prompt} \\
\midrule
Counting & Edit & \texttt{\{new\_total\_legs\} leg(s) total. Edit this image so the \{animal\_name\} has exactly \{new\_total\_legs\} leg(s). \{arrangement hint\}. Every leg must match the size and texture of the original legs. The result must be photo-realistic with all \{new\_total\_legs\} leg(s) clearly visible and distinct. Preserve the same background, facial features, lighting, and camera angle.} \\
\addlinespace
Direction & Edit & \texttt{Edit this image so the \{object\_name\} faces \{target\_direction\}ward. \{front\_part\} should point toward the \{top|bottom\} of the frame, with the \{object\_name\} \{pose\_desc\}. The \{object\_name\} must remain clearly visible and recognizable. The result must be photo-realistic. Preserve the same background, lighting, object size, and camera angle.} \\
\addlinespace
Spatial & Caption & \texttt{You generate counterfactual captions by changing ONLY the spatial relation between the same two objects. Caption: \{original\_caption\}. Choose a preposition from \{allowed\_preps\} that makes the resulting caption feel the most unnatural. Produce \{num\_variants\} candidate captions.} \\
\addlinespace
Spatial & Edit & \texttt{Edit this image. From the viewer's perspective, the \{obj1\} is \{relation\} the \{obj2\}. Both objects must remain fully visible. Maintain photorealistic appearance. Preserve the same background, lighting, and object appearance.} \\
\addlinespace
Color ID & Generate & \texttt{Generate a photorealistic image of a \{animal\_name\}. The \{animal\_name\} should be the central object in the image. Include realistic natural scenery around the animal. The animal must be clearly visible.} \\
\addlinespace
Color ID & Edit & \texttt{Edit this image: change the color of the \{animal\_name\} from \{original\_color\} to \{target\_color\}. The \{animal\_name\} must remain clearly recognizable. Preserve the same background, lighting, and camera angle.} \\
\addlinespace
Shape ID & Generate & \texttt{Generate a photorealistic image of a \{sign\_name\} traffic sign. The sign is \{shape\}-shaped and should be the central object in the image. Include realistic outdoor scenery. The sign text and markings must be clearly visible.} \\
\addlinespace
Shape ID & Edit & \texttt{Edit this image: change the shape of the \{sign\_name\} sign from \{original\_shape\} to \{target\_shape\}. Keep the sign's text, colors, and markings identical. The \{target\_shape\}-shaped sign must be clearly visible. Preserve the same background, lighting, and camera angle.} \\
\bottomrule
\end{tabular}
\end{table}

\subsubsection{Evaluation and Models}

\paragraph{Prompt Setup for In-Context Learning.}
\label{app:evaluation_settings}
Each demonstration is a triple of an image $I$, a question $Q$, and an answer $A$. The full prompt begins with a system prefix, followed by $N$ demonstrations $(I_1, Q_1, A_1), \dots, (I_N, Q_N, A_N)$ presented as interleaved image--answer pairs, and ends with the query pair $(I, Q)$ whose answer the model must predict.

\subparagraph{Demonstration sampling.}
We use $N \in \{0, 1, 2, 4, 8, 16, 32\}$ demonstrations, where $N{=}0$ is the zero-shot baseline. Demonstrations are drawn uniformly across the unique answer labels in two steps. Let $L$ be the number of unique labels. First, we sample $\lfloor N / L \rfloor$ demonstrations from each label so that every label is equally represented. The remaining $N \bmod L$ slots are then filled by sampling uniformly from the leftover pool across all labels. This label-balanced sampling prevents improvements from being explained by answer-label frequency in the demonstrations.
For \emph{complex} \settings, each context also balances real and counterfactual demonstrations whenever the shot budget and candidate pool allow.

Demonstrations and queries are drawn from the same task and \setting unless otherwise specified (e.g., experiments across \settings in Section~\ref{sec:performance_evaluation}).
Query images are strictly disjoint from the demonstration candidate pool.
Each nonzero-shot query is evaluated under 10 fixed demonstration contexts sampled with this procedure and a fixed random seed.
Accuracy is averaged over query--context pairs, so each query contributes equally after averaging over its contexts.
When an analysis pools tasks or \settings, we report the micro-average over evaluated query entries rather than an unweighted mean of task-level accuracies.
We use the same contexts across model families and prompt conditions whenever applicable.
These repeated contexts reduce sensitivity to which specific demonstrations a query is paired with, since some demonstration sets may align unusually well or poorly with a given query.

For object direction, the real/counterfactual query split is also tied to orientation labels: real queries use EgoOrientBench's front, back, left, and right directions, while edited counterfactual queries introduce up and down.
We therefore interpret this task as a rare-orientation stress test rather than a fully label-matched real/counterfactual comparison.
The ICL contexts remain label-balanced across all six directions; at 16 shots, each context contains at least two examples per direction, including up and down, with real and counterfactual examples mixed.

\paragraph{Output Scoring and Answer Distributions.}
\label{app:output_scoring}

For exact-match scoring, we first extract the text inside the last pair of curly braces, \texttt{\{...\}}, in the model output.
We then lowercase this text and strip leading and trailing whitespace.
We apply the same lowercase-and-strip normalization to the ground-truth answer.
The prediction is correct only if the two normalized strings match exactly.
If the model output has no curly-braced span, the prediction is marked incorrect.

For answer-distribution analyses in Appendix~\ref{app:answer_label_probability}, we score candidate labels using normalized token-logit probabilities. For each query, let $\mathcal{Y}$ be the task-specific candidate label set. We compute a normalized probability for each candidate answer by applying a softmax over the candidate label logits. We then report the probability assigned to the visually correct answer $P(y_{\mathrm{visual}})$ and the prior/default answer $P(y_{\mathrm{prior}})$. This measurement is separate from raw answer frequency over sampled demonstration sets.

\paragraph{Model Architectures and Configurations.}
\label{app:model_settings}
We evaluate five model families spanning a wide range of parameter counts, covering both recent vision--language models and earlier generations used as baselines. The specific checkpoints are:
\begin{itemize}
    \item \textbf{Qwen3-VL}~\citep{bai2025qwen3vltechnicalreport}: 2B, 4B, 8B, 32B, and 235B parameters.
    \item \textbf{Qwen2.5-VL}~\citep{qwen2025qwen25technicalreport}: 3B and 32B parameters.
    \item \textbf{Qwen3.5}~\citep{qwen35blog}: 4B and 27B parameters.
    \item \textbf{InternVL3.5}~\citep{wang2025internvl35advancingopensourcemultimodal}: 4B and 38B parameters.
    \item \textbf{Gemma-4}~\citep{gemma4_2026}: E4B and 31B parameters.
\end{itemize}
For each family we use the instruction-tuned checkpoint.
For models that support an extended ``thinking'' mode, we disable it by default so that all comparisons use single-pass, non-reasoning inference.
Results with thinking enabled are reported separately in Section~\ref{subsec:reasoning_icl}.
Main behavioral evaluations use vLLM~\citep{kwon2023efficientmemorymanagementlarge}.
Mechanistic evaluations use Hugging Face Transformers.
For benchmark evaluations, we use the sampling parameters recommended by each model's official model card (summarized in Table~\ref{tab:app_sampling_params}).
For mechanistic experiments, we use greedy decoding so that internal probes (attention, hidden states, label-logit probabilities) are evaluated against a deterministic generation rather than a stochastic sample.

\begin{table}[h]
\centering
\caption{\textbf{Sampling parameters for VLM benchmark evaluation.} Values follow each model card's recommended decoding settings and are used for the vLLM behavioral runs. Mechanistic experiments (Hugging Face Transformers) instead use greedy decoding and therefore do not use the temperature, top-$p$, top-$k$, repetition, or presence parameters listed here.}
\label{tab:app_sampling_params}
\scriptsize
\setlength{\tabcolsep}{4pt}
\renewcommand{\arraystretch}{1.05}
\resizebox{\textwidth}{!}{%
\begin{tabular}{llcccccc}
\toprule
\textbf{Mode} & \textbf{Model family} & \textbf{Max tokens} & \textbf{Temp.} & \textbf{Top-$p$} & \textbf{Top-$k$} & \textbf{Rep. penalty} & \textbf{Presence penalty} \\
\midrule
Standard non-thinking & Qwen3-VL, Qwen2.5-VL, Qwen3.5, InternVL3.5 & $512$ & $0.7$ & $0.8$ & $20$ & $1.0$ & vLLM: $1.5$; HF: omitted \\
Standard non-thinking & Gemma-4 & $512$ & $1.0$ & $0.95$ & $64$ & $1.0$ & vLLM: $0.0$; HF: omitted \\
\midrule
Thinking mode & Qwen3-VL, Qwen2.5-VL, Qwen3.5, InternVL3.5 & $1024$ & $1.0$ & $0.95$ & $20$ & N/A & N/A \\
Thinking mode & Gemma-4 & $1024$ & $1.0$ & $0.95$ & $64$ & N/A & N/A \\
\bottomrule
\end{tabular}
}
\end{table}

All images are resized to $512 \times 512$ pixels before inference.
We use each model's default system prompt and default chat-template formatting.
There are no task-specific prompt differences across model families beyond backend-required serialization.

\paragraph{Compute Resources.}
\label{app:compute}
All experiments are run on a single node with 4 NVIDIA H200 GPUs.
Behavioral ICL evaluations use vLLM~\citep{kwon2023efficientmemorymanagementlarge} for batched multi-image inference, and mechanistic analyses use Hugging Face Transformers to access per-layer attention and hidden states.
The full empirical ICL sweep, covering all five model families, both representative model sizes per family, all five tasks, all three \settings, and shot counts $N \in \{0, 1, 2, 4, 8, 16, 32\}$ with 10 demonstration contexts per task, \setting, and shot count, consumed approximately 400 GPU-hours in aggregate.
Mechanistic experiments and preliminary or discarded runs are not included in this estimate.

\subsection{Further Evaluations}
\label{app:further_evaluations}
\label{app:TODO}

\subsubsection{ICL Effect Across Model Scales}
This section extends the scaling results presented in Section~\ref{sec:performance_evaluation} with a more comprehensive analysis. 
We evaluate ICL performance across five model families: Qwen3-VL~\citep{bai2025qwen3vltechnicalreport}, Qwen2.5-VL~\citep{qwen2025qwen25technicalreport}, Qwen3.5~\citep{qwen35blog}, InternVL3.5~\citep{wang2025internvl35advancingopensourcemultimodal}, and Gemma-4~\citep{gemma4_2026}.
For each family, we select two representative sizes—4B and 30B—to examine how parameter count influences scaling behavior.

\begin{figure}[t]
    \centering
    \includegraphics[width=\linewidth]{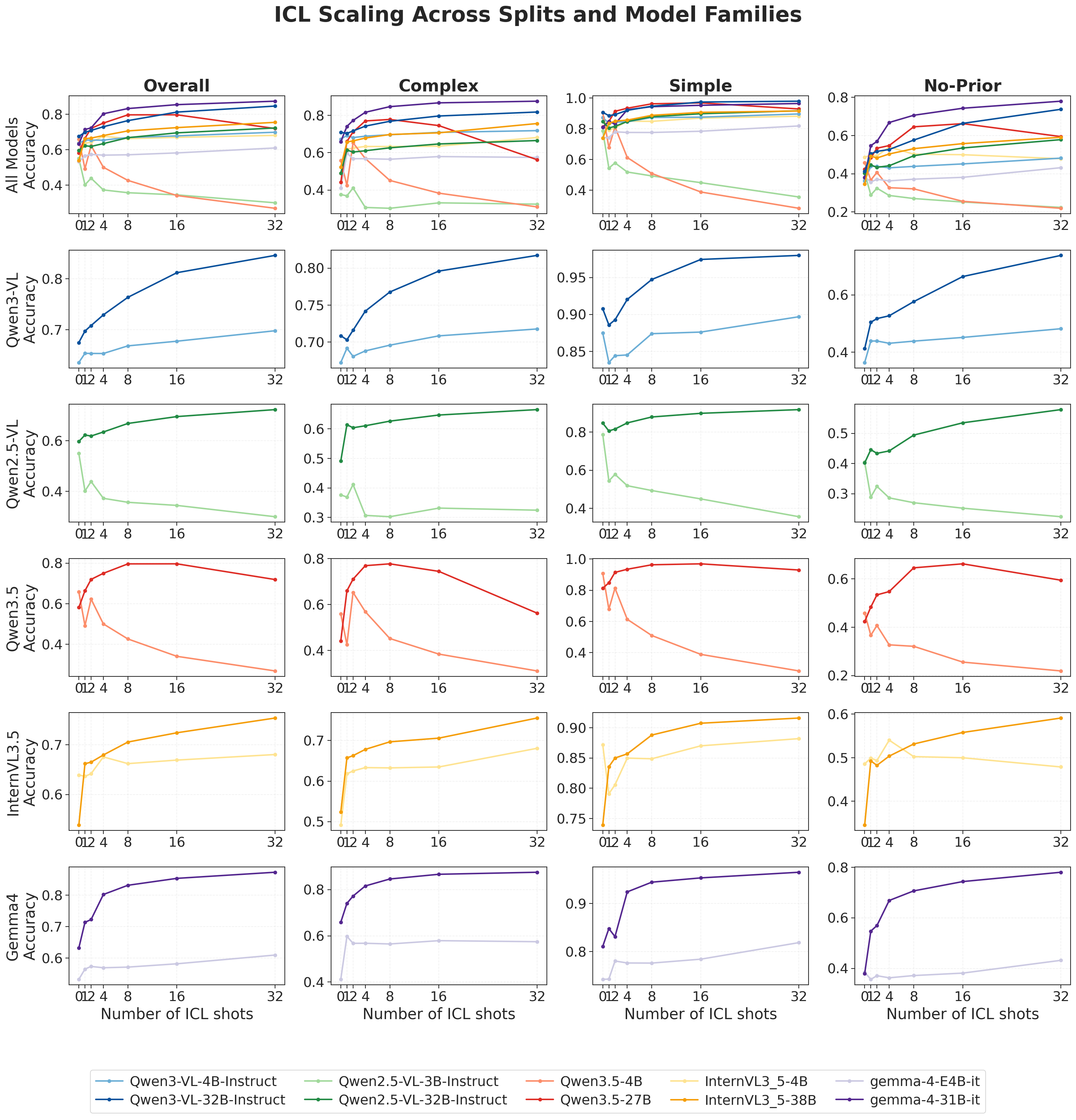}
    \caption{\textbf{ICL scaling across model families and \settings.} Rows: model families (aggregated, then Qwen3-VL, Qwen2.5-VL, Qwen3.5, InternVL3.5, Gemma-4). Columns: \settings. Qwen3-VL scales consistently; Qwen2.5-VL less so. Qwen3.5 and InternVL3.5 degrade beyond $\sim$8--16 shots. Gemma-4 achieves the strongest and most stable scaling.}
    \label{fig:app_modelfamily_scaling}
\end{figure}

Figure~\ref{fig:app_modelfamily_scaling} reports performance overall and across the \emph{complex}, \emph{simple}, and \emph{no-prior} \settings of the dataset. 
Two trends emerge. First, performance generally improves as the number of demonstrations grows, although Qwen2.5-VL 3B and Qwen3.5 show non-monotonic scaling.
Second, larger models benefit more from additional demonstrations than their smaller counterparts.

\paragraph{Special Cases: Non-Monotonic Scaling}
On \emph{complex} examples, Qwen2.5-VL 3B falls below zero-shot accuracy at 4 and 16 shots.
Qwen3.5 4B improves at 4 shots, then declines on both real and counterfactual examples at 16; 27B retains most gains (Figure~\ref{fig:app_modelfamily_scaling}).
One possible cause is long-context degradation: 16 or 32 multimodal demonstrations can push the context above 100K tokens.

\subsubsection{Task Performance Breakdown in Detail}
\label{app:task_breakdown}

Table~\ref{tab:app_cross_family_accuracies} reports task-level accuracies for each evaluated model family and \setting, showing performance at 0-shot, 16-shot, and the difference between the two when both endpoints are available. 
For each family, we select two representative sizes—4B and 30B—to examine how parameter count influences scaling behavior.

Rows with both endpoints available show that the \emph{complex} \setting often gains from in-context demonstrations, although the size and direction of the gain vary by task and model. 
Gemma-4 is the strongest performer overall, achieving both the highest absolute accuracies and the largest gains from demonstrations on most tasks and \settings. 
Within most model families, larger models generally benefit more from demonstrations, although task-level gains are not monotonic across all tasks and \settings.
% Cross-family accuracies; values are percentages rounded to one decimal place.
% Nonzero-shot cells include standard deviations over sampled demonstration contexts.
% Each task block reports 0-shot, 16-shot, and $\Delta_{16-0}$ for three settings.

\providecommand{\taskperfstd}[1]{{\tiny\,($\pm#1$)}}

\begin{table}[!htbp]
    \centering
    \caption{\textbf{Cross-family accuracies by task and \setting.} Accuracies are percentages; standard deviations and $\Delta_{16-0}$ are in percentage points. For 16-shot rows, parentheses report standard deviations over sampled demonstration contexts. Best 0- and 16-shot entries per row are bold; for $\Delta_{16-0}$ the largest per-row improvement is bold. A dash indicates that a valid run was unavailable, and $\Delta_{16-0}$ is omitted unless both endpoints are available.}
    \label{tab:app_cross_family_accuracies}
    \scriptsize
    \setlength{\tabcolsep}{4.5pt}
    \renewcommand{\arraystretch}{1}
    \resizebox{\linewidth}{!}{%
    \begin{tabular}{llrrrrrrrrrr}
    \toprule
    & & \multicolumn{2}{c}{\textbf{InternVL3.5}} & \multicolumn{2}{c}{\textbf{Qwen2.5-VL}} & \multicolumn{2}{c}{\textbf{Qwen3-VL}} & \multicolumn{2}{c}{\textbf{Qwen3.5}} & \multicolumn{2}{c}{\textbf{Gemma-4}} \\
    \cmidrule(lr){3-4}\cmidrule(lr){5-6}\cmidrule(lr){7-8}\cmidrule(lr){9-10}\cmidrule(l){11-12}
    \textbf{\Setting} & \textbf{Shots} & \textbf{4B} & \textbf{38B} & \textbf{3B} & \textbf{32B} & \textbf{4B} & \textbf{32B} & \textbf{4B} & \textbf{27B} & \textbf{E4B} & \textbf{31B} \\
    \midrule
    \multicolumn{12}{l}{\textit{Counting}} \\
    \midrule
    \multirow{3}{*}{No-prior} & 0 & 43.1 & 28.8 & 35.6 & 41.9 & 32.5 & 43.8 & \textbf{58.8} & 49.7 & 35.0 & 23.8 \\
     & 16 & 49.8\taskperfstd{29.4} & 76.5\taskperfstd{26.3} & 19.9\taskperfstd{25.9} & 70.8\taskperfstd{20.0} & 35.6\taskperfstd{18.1} & 71.9\taskperfstd{18.6} & 16.8\taskperfstd{30.0} & \textbf{80.4}\taskperfstd{22.6} & 28.6\taskperfstd{22.5} & 72.4\taskperfstd{25.8} \\
     & $\Delta_{16-0}$ & +6.6 & +47.8 & -15.7 & +28.9 & +3.1 & +28.2 & -41.9 & +30.8 & -6.4 & \textbf{+48.5} \\
    \cmidrule(l){1-12}
    \multirow{3}{*}{Simple} & 0 & 92.5 & 63.8 & 87.5 & 84.4 & 88.1 & 93.8 & 95.0 & \textbf{95.6} & 63.8 & 54.4 \\
     & 16 & 85.2\taskperfstd{18.0} & 89.1\taskperfstd{10.2} & 26.7\taskperfstd{23.8} & 82.9\taskperfstd{9.0} & 82.1\taskperfstd{13.2} & \textbf{95.5}\taskperfstd{3.6} & 23.3\taskperfstd{33.5} & 94.6\taskperfstd{8.8} & 65.6\taskperfstd{29.0} & 83.8\taskperfstd{17.6} \\
     & $\Delta_{16-0}$ & -7.2 & +25.3 & -60.8 & -1.4 & -6.1 & +1.8 & -71.7 & -1.1 & +1.8 & \textbf{+29.4} \\
    \cmidrule(l){1-12}
    \multirow{3}{*}{Complex} & 0 & 30.0 & 30.0 & 31.0 & 30.0 & 30.0 & 33.0 & 30.0 & 30.0 & 30.0 & \textbf{42.0} \\
     & 16 & 30.7\taskperfstd{26.7} & 30.6\taskperfstd{3.2} & 26.1\taskperfstd{15.0} & 31.0\taskperfstd{1.9} & 36.0\taskperfstd{12.5} & 53.7\taskperfstd{14.6} & 25.9\taskperfstd{16.8} & 46.6\taskperfstd{18.4} & 29.8\taskperfstd{4.5} & \textbf{76.4}\taskperfstd{15.8} \\
     & $\Delta_{16-0}$ & +0.7 & +0.6 & -4.9 & +1.0 & +6.0 & +20.7 & -4.1 & +16.6 & -0.2 & \textbf{+34.4} \\
    \midrule
    \multicolumn{12}{l}{\textit{Color identification}} \\
    \midrule
    \multirow{3}{*}{No-prior} & 0 & \textbf{85.0} & 50.8 & 65.0 & 52.5 & 50.8 & 52.5 & 61.7 & 60.0 & 55.8 & 54.2 \\
     & 16 & \textbf{84.8}\taskperfstd{17.0} & 70.4\taskperfstd{19.9} & 37.9\taskperfstd{34.3} & 78.6\taskperfstd{18.3} & 63.4\taskperfstd{17.2} & 84.5\taskperfstd{17.2} & 38.8\taskperfstd{33.5} & 72.0\taskperfstd{30.9} & 53.8\taskperfstd{17.6} & 81.2\taskperfstd{20.6} \\
     & $\Delta_{16-0}$ & -0.2 & +19.6 & -27.1 & +26.1 & +12.6 & \textbf{+32.0} & -22.8 & +12.0 & -2.1 & +27.1 \\
    \cmidrule(l){1-12}
    \multirow{3}{*}{Simple} & 0 & \textbf{100.0} & \textbf{100.0} & 98.3 & \textbf{100.0} & \textbf{100.0} & \textbf{100.0} & \textbf{100.0} & \textbf{100.0} & \textbf{100.0} & \textbf{100.0} \\
     & 16 & 99.8\taskperfstd{0.4} & \textbf{100.0}\taskperfstd{0.0} & 79.1\taskperfstd{25.9} & \textbf{100.0}\taskperfstd{0.0} & 99.6\taskperfstd{0.7} & \textbf{100.0}\taskperfstd{0.0} & 58.2\taskperfstd{45.0} & 99.4\taskperfstd{1.8} & 99.4\taskperfstd{1.3} & 99.9\taskperfstd{0.3} \\
     & $\Delta_{16-0}$ & -0.2 & \textbf{+0.0} & -19.2 & \textbf{+0.0} & -0.4 & \textbf{+0.0} & -41.8 & -0.6 & -0.6 & -0.1 \\
    \cmidrule(l){1-12}
    \multirow{3}{*}{Complex} & 0 & 83.5 & 83.7 & 47.2 & 79.1 & \textbf{86.7} & 80.8 & 56.4 & 78.8 & 77.9 & 84.3 \\
     & 16 & 87.9\taskperfstd{4.6} & 86.3\taskperfstd{8.5} & 46.2\taskperfstd{41.9} & 82.9\taskperfstd{9.0} & 88.2\taskperfstd{6.5} & 92.3\taskperfstd{6.5} & 48.8\taskperfstd{40.8} & 92.2\taskperfstd{8.9} & 77.8\taskperfstd{20.3} & \textbf{93.9}\taskperfstd{5.5} \\
     & $\Delta_{16-0}$ & +4.5 & +2.7 & -1.0 & +3.8 & +1.5 & +11.5 & -7.6 & \textbf{+13.4} & -0.1 & +9.6 \\
    \midrule
    \multicolumn{12}{l}{\textit{Shape identification}} \\
    \midrule
    \multirow{3}{*}{No-prior} & 0 & 63.3 & 60.0 & 59.2 & 59.2 & 55.8 & 61.7 & 60.0 & 57.5 & 58.8 & \textbf{67.9} \\
     & 16 & 65.9\taskperfstd{16.1} & 58.0\taskperfstd{19.6} & 39.7\taskperfstd{23.7} & 56.0\taskperfstd{14.0} & 68.2\taskperfstd{21.1} & 64.2\taskperfstd{19.6} & 38.2\taskperfstd{43.7} & 59.5\taskperfstd{26.3} & 61.2\taskperfstd{19.4} & \textbf{81.1}\taskperfstd{23.7} \\
     & $\Delta_{16-0}$ & +2.6 & -2.0 & -19.5 & -3.2 & +12.4 & +2.6 & -21.8 & +2.0 & +2.4 & \textbf{+13.2} \\
    \cmidrule(l){1-12}
    \multirow{3}{*}{Simple} & 0 & 99.2 & 98.3 & 98.3 & 98.3 & \textbf{100.0} & \textbf{100.0} & \textbf{100.0} & \textbf{100.0} & 90.8 & \textbf{100.0} \\
     & 16 & 99.2\taskperfstd{0.9} & 99.8\taskperfstd{0.8} & 54.5\taskperfstd{39.1} & \textbf{100.0}\taskperfstd{0.0} & 99.9\taskperfstd{0.3} & \textbf{100.0}\taskperfstd{0.0} & 49.8\taskperfstd{47.2} & 97.2\taskperfstd{7.0} & 94.9\taskperfstd{8.2} & 98.9\taskperfstd{2.6} \\
     & $\Delta_{16-0}$ & +0.1 & +1.4 & -43.8 & +1.7 & -0.1 & +0.0 & -50.2 & -2.8 & \textbf{+4.1} & -1.1 \\
    \cmidrule(l){1-12}
    \multirow{3}{*}{Complex} & 0 & 83.0 & 93.7 & 48.7 & 83.2 & \textbf{94.2} & 87.5 & 22.7 & 82.3 & 89.5 & \textbf{94.2} \\
     & 16 & 85.4\taskperfstd{5.7} & 96.6\taskperfstd{3.4} & 27.0\taskperfstd{18.3} & 87.6\taskperfstd{9.6} & 97.8\taskperfstd{3.7} & 97.3\taskperfstd{2.0} & 29.2\taskperfstd{42.0} & 98.9\taskperfstd{2.2} & 89.2\taskperfstd{13.7} & \textbf{99.9}\taskperfstd{0.3} \\
     & $\Delta_{16-0}$ & +2.4 & +2.9 & -21.7 & +4.4 & +3.6 & +9.8 & +6.6 & \textbf{+16.7} & -0.3 & +5.8 \\
    \midrule
    \multicolumn{12}{l}{\textit{Direction}} \\
    \midrule
    \multirow{3}{*}{No-prior} & 0 & \textbf{26.7} & 15.0 & 16.7 & 25.0 & 20.0 & 22.5 & 19.7 & 14.2 & 17.2 & 15.8 \\
     & 16 & 26.0\taskperfstd{26.5} & 37.9\taskperfstd{23.5} & 18.9\taskperfstd{9.0} & 34.2\taskperfstd{28.2} & 37.9\taskperfstd{34.1} & 71.4\taskperfstd{28.4} & 19.7\taskperfstd{28.1} & 73.8\taskperfstd{32.9} & 28.8\taskperfstd{16.3} & \textbf{92.0}\taskperfstd{13.3} \\
     & $\Delta_{16-0}$ & -0.7 & +22.9 & +2.2 & +9.2 & +17.9 & +48.9 & +0.1 & +59.7 & +11.6 & \textbf{+76.2} \\
    \cmidrule(l){1-12}
    \multirow{3}{*}{Simple} & 0 & 68.3 & 50.0 & 43.3 & 70.0 & 77.5 & 78.2 & 72.5 & 15.0 & 55.0 & \textbf{79.2} \\
     & 16 & 79.4\taskperfstd{20.3} & 89.2\taskperfstd{14.3} & 50.4\taskperfstd{28.1} & 89.4\taskperfstd{8.6} & 83.1\taskperfstd{6.3} & 99.7\taskperfstd{1.1} & 33.2\taskperfstd{38.9} & 99.9\taskperfstd{0.3} & 75.1\taskperfstd{17.0} & \textbf{100.0}\taskperfstd{0.0} \\
     & $\Delta_{16-0}$ & +11.1 & +39.2 & +7.1 & +19.4 & +5.6 & +21.5 & -39.3 & \textbf{+84.9} & +20.0 & +20.8 \\
    \cmidrule(l){1-12}
    \multirow{3}{*}{Complex} & 0 & 56.7 & 70.0 & 34.2 & 52.5 & 58.3 & \textbf{77.5} & 64.2 & 28.3 & 38.3 & \textbf{77.5} \\
     & 16 & 56.8\taskperfstd{19.7} & 73.2\taskperfstd{11.9} & 29.2\taskperfstd{30.2} & 55.8\taskperfstd{8.9} & 65.6\taskperfstd{11.1} & 84.2\taskperfstd{8.5} & 22.3\taskperfstd{37.9} & 72.8\taskperfstd{23.3} & 44.0\taskperfstd{18.4} & \textbf{87.8}\taskperfstd{8.5} \\
     & $\Delta_{16-0}$ & +0.2 & +3.2 & -5.0 & +3.2 & +7.2 & +6.7 & -41.8 & \textbf{+44.5} & +5.7 & +10.3 \\
    \midrule
    \multicolumn{12}{l}{\textit{Spatial relation}} \\
    \midrule
    \multirow{3}{*}{No-prior} & 0 & 26.7 & 20.0 & 27.5 & 21.9 & 24.2 & 25.0 & 25.0 & 28.3 & 24.5 & \textbf{33.3} \\
     & 16 & 23.6\taskperfstd{6.0} & 29.2\taskperfstd{12.3} & 11.4\taskperfstd{18.9} & 22.2\taskperfstd{14.4} & 23.8\taskperfstd{3.1} & 38.0\taskperfstd{17.2} & 17.1\taskperfstd{28.4} & 40.8\taskperfstd{27.6} & 21.4\taskperfstd{18.0} & \textbf{45.6}\taskperfstd{25.0} \\
     & $\Delta_{16-0}$ & -3.1 & +9.2 & -16.1 & +0.4 & -0.3 & \textbf{+13.0} & -7.9 & +12.4 & -3.1 & +12.2 \\
    \cmidrule(l){1-12}
    \multirow{3}{*}{Simple} & 0 & 74.2 & 60.8 & 63.3 & 70.8 & 71.7 & 80.8 & 85.0 & \textbf{90.8} & 63.3 & 80.8 \\
     & 16 & 71.9\taskperfstd{7.2} & 76.3\taskperfstd{8.1} & 20.2\taskperfstd{32.3} & 79.1\taskperfstd{8.2} & 75.2\taskperfstd{6.5} & 92.8\taskperfstd{4.0} & 35.2\taskperfstd{39.3} & 94.2\taskperfstd{7.3} & 61.5\taskperfstd{11.3} & \textbf{97.8}\taskperfstd{1.7} \\
     & $\Delta_{16-0}$ & -2.2 & +15.5 & -43.2 & +8.2 & +3.6 & +11.9 & -49.8 & +3.4 & -1.8 & \textbf{+17.0} \\
    \cmidrule(l){1-12}
    \multirow{3}{*}{Complex} & 0 & 58.3 & 53.3 & 46.7 & 61.7 & 60.8 & 69.2 & 69.2 & 71.7 & 53.3 & \textbf{74.2} \\
     & 16 & 51.1\taskperfstd{7.8} & 59.6\taskperfstd{9.1} & 36.2\taskperfstd{8.8} & 60.4\taskperfstd{7.5} & 60.9\taskperfstd{6.1} & 72.2\taskperfstd{4.9} & 64.4\taskperfstd{17.5} & \textbf{78.2}\taskperfstd{11.3} & 44.1\taskperfstd{16.2} & 73.4\taskperfstd{9.4} \\
     & $\Delta_{16-0}$ & -7.2 & +6.2 & -10.4 & -1.2 & +0.1 & +3.0 & -4.8 & \textbf{+6.5} & -9.2 & -0.8 \\
    \bottomrule
    \end{tabular}
    }
\end{table}

\paragraph{On the Real vs.\ Counterfactual Asymmetry.}
A natural concern about Figure~\ref{fig:accuracies_cf}(b) is that real-image accuracy on color and shape identification already exceeds $95.0\%$ at zero shot for most $\geq$30B models, mechanically capping the absolute real-side $\Delta$ at only a few percentage points.
The headline finding does not depend on these two tasks. 
Aggregate real accuracy on the \emph{complex} \setting is $86.9\%$ at zero shot for Qwen3-VL 32B (Section~\ref{sec:performance_evaluation}), leaving roughly $13.0$ percentage points of headroom that 16-shot prompting does not consume on the real subset.
Counting, object direction, and spatial relation all have real-image zero-shot accuracy well below ceiling (Table~\ref{tab:app_cross_family_accuracies}), and the cf-vs-real asymmetry holds on these tasks as well. 

Headroom-normalized error reduction, $\mathrm{ERR}=(\mathrm{Acc}_{16}-\mathrm{Acc}_{0})/(1-\mathrm{Acc}_{0})$, measures the fraction of zero-shot error removed by demonstrations, using fractional accuracies.
For Qwen3-VL 32B on \emph{complex} color identification, real-image accuracy falls from $91.7\%$ to $87.9\%$, while counterfactual accuracy rises from $70.0\%$ to $96.7\%$.
The corresponding ERR values are $-45.8\%$ and $+89.0\%$: demonstrations increase real-image error by $45.8\%$ but reduce counterfactual error by $89.0\%$.
Thus, the counterfactual advantage persists after accounting for the different amounts of available headroom.

\paragraph{Paired Errors of Gemma-4 and Qwen3-VL.}
\label{app:paired_errors}
To locate where the higher accuracy of Gemma-4 31B comes from, we compare it with Qwen3-VL 32B on the same \emph{complex} queries (Table~\ref{tab:paired_errors}).
On real examples, agreement stays stable at $89.6\%$ at 0 shots and $88.9\%$ at 16 shots.
On counterfactual examples, 16-shot prompting raises shared correctness from $51.7\%$ to $72.6\%$ and lowers shared errors from $23.7\%$ to $10.4\%$, so agreement increases by $7.7$ points as the two models converge on the grounded answer.
Gemma-4-only successes exceed Qwen3-VL-only successes in every row, and the gap is widest on counterfactual examples ($15.4\%$ versus $9.2\%$ at 0 shots, $13.5\%$ versus $3.4\%$ at 16 shots).
Counting accounts for much of this advantage: on counterfactual counting queries, Gemma-4 31B answers $21.3\%$ correctly at 0 shots versus $4.2\%$ for Qwen3-VL 32B, and $70.1\%$ versus $35.8\%$ at 16 shots.
Overall counting accuracy is $42.0\%$ versus $33.0\%$ at 0 shots and $76.4\%$ versus $53.7\%$ at 16 shots, the largest per-task gap in Table~\ref{tab:app_cross_family_accuracies}.
Demonstrations therefore move both models toward the grounded answer on examples where the \prior conflicts with the image, not only the weaker model.

% Source: rebuttal/R1D.md, Part 2, Table 1, which R1D labels "Estimated".
% The Qwen3-VL 32B marginals (both correct + Qwen-only) match the Figure 4(a)
% real and counterfactual values.
% The per-task paired table in R1D (Table 2) is omitted because its counting
% rows violate the per-model accuracy bounds in rebuttal/R1R.md, Table R1.7.
% The counting accuracies cited in the text come from R1R Table R1.7.
\begin{table}[!htbp]
  \centering
  \caption{\textbf{Paired correctness of Gemma-4 31B and Qwen3-VL 32B.} Both models answer the same \emph{complex} queries, split into real and counterfactual examples, at 0 and 16 shots. Agreement is the share of cases where both models are correct or both are incorrect. Percentages use the same aggregation as Figure~\ref{fig:accuracies_cf}(a), and entries are rounded independently.}
  \label{tab:paired_errors}
  \small
  \setlength{\tabcolsep}{4pt}
  \resizebox{\linewidth}{!}{%
  \begin{tabular}{@{}llrrrrrr@{}}
    \toprule
    \textbf{Subset} & \textbf{Shots} & \textbf{Both correct (\%)} & \textbf{Gemma-4 only (\%)} & \textbf{Qwen3-VL only (\%)} & \textbf{Both incorrect (\%)} & \textbf{Agreement (\%)} & \textbf{$\Delta$ Agreement (pts)} \\
    \midrule
    \multirow{2}{*}{Real}           & 0  & 82.4 & 5.9  & 4.5 & 7.2  & 89.6 & --- \\
                                    & 16 & 80.9 & 6.6  & 4.5 & 8.1  & 88.9 & $-$0.7 \\
    \midrule
    \multirow{2}{*}{Counterfactual} & 0  & 51.7 & 15.4 & 9.2 & 23.7 & 75.4 & --- \\
                                    & 16 & 72.6 & 13.5 & 3.4 & 10.4 & 83.1 & +7.7 \\
    \bottomrule
  \end{tabular}%
  }
\end{table}

\subsubsection{Expanded Benchmark Evaluation}
\label{app:expanded_benchmark}

We expand each task to approximately 400 complex examples by adding only counterfactual images.
This changes the real-example share from $56.0\%$ to approximately $33\%$; exact expanded per-task counts are unavailable.
Table~\ref{tab:expanded_benchmark} reports a partial evaluation with Qwen3-VL 32B at 0 and 16 shots.
Accuracy improves with 16 demonstrations on all five tasks, although these results cover only one model and two shot counts.
Because overall task accuracy weights the real and counterfactual subsets by their counts, comparisons with the original benchmark also reflect the change in subset composition.

% Source: rebuttal/R1R.md, Sections 2--3 and Table R1.4.
% Deltas are differences of the reported, rounded means.
\begin{table}[!htbp]
  \centering
  \small
  \begin{subtable}[t]{0.88\linewidth}
    \centering
    \setlength{\tabcolsep}{5pt}
    \begin{tabular*}{\linewidth}{@{}l@{\extracolsep{\fill}}cccc@{}}
      \toprule
      \textbf{Version} & \textbf{Complex Examples} & \textbf{Real} & \textbf{Counterfactual} & \textbf{Real Share} \\
      \midrule
      Original & 1,160 total & 650 & 510 & $56.0\%$ \\
      Expanded & $\sim$400 per task & 650 & not reported & $\sim33\%$ \\
      \bottomrule
    \end{tabular*}
    \caption{Dataset statistics}
  \end{subtable}

  \medskip
  \begin{subtable}[t]{0.88\linewidth}
    \centering
    \setlength{\tabcolsep}{7pt}
    \begin{tabular*}{\linewidth}{@{}l@{\extracolsep{\fill}}rrr@{}}
      \toprule
      \textbf{Task} & \textbf{0-shot (\%)} & \textbf{16-shot (\%)} & $\boldsymbol{\Delta}$ \textbf{(pp)} \\
      \midrule
      Counting         & 23.5 & $48.3 \pm 12.1$ & +24.8 \\
      Object Direction & 65.2 & $79.5 \pm 12.8$ & +14.3 \\
      Spatial Relation & 62.8 & $70.0 \pm 7.2$ & +7.2 \\
      Color ID         & 72.9 & $89.1 \pm 8.9$ & +16.2 \\
      Shape ID         & 83.2 & $96.4 \pm 3.1$ & +13.2 \\
      \bottomrule
    \end{tabular*}
    \caption{Partial expanded-benchmark results}
  \end{subtable}
  \caption{Expanded complex-setting dataset and partial results. Exact expanded counterfactual and total counts are unavailable. The partial evaluation covers Qwen3-VL 32B at two shot counts; 16-shot entries are mean accuracy $\pm$ standard deviation across ten fixed demonstration contexts, while zero-shot has no context variance. Accuracies are percentages; standard deviations and $\Delta$ are in percentage points (pp).}
  \label{tab:expanded_benchmark}
\end{table}

\subsubsection{Masking and Length-Matched Control Ablations}
\label{app:unlearning}

Table~\ref{tab:control_ablations} tests whether the gains come from added context, prompt format, or indiscriminate token removal.
These control ablations use Qwen3-VL 32B on the \emph{complex} \setting with greedy decoding unless noted in the table.
At 16 shots, masking grounding cues in the query image reduces accuracy by between $20.7$ and $30.1$ percentage points across counting, color identification, and shape identification.
By contrast, prior-cue masking on real images changes zero-shot accuracy by at most $4.7$ points.
This contrast is consistent with demonstrations steering the model toward grounding cues rather than providing a generic masking benefit.

The length-matched controls separate demonstration content from context length.
Blank images reduce all three tasks below zero-shot accuracy.
On counting, cross-split counting and vision-only demonstrations recover part of the full 16-shot gain, but both remain below the standard demonstrations.

\begin{figure}[t]
    \centering
    \includegraphics[width=0.60\linewidth]{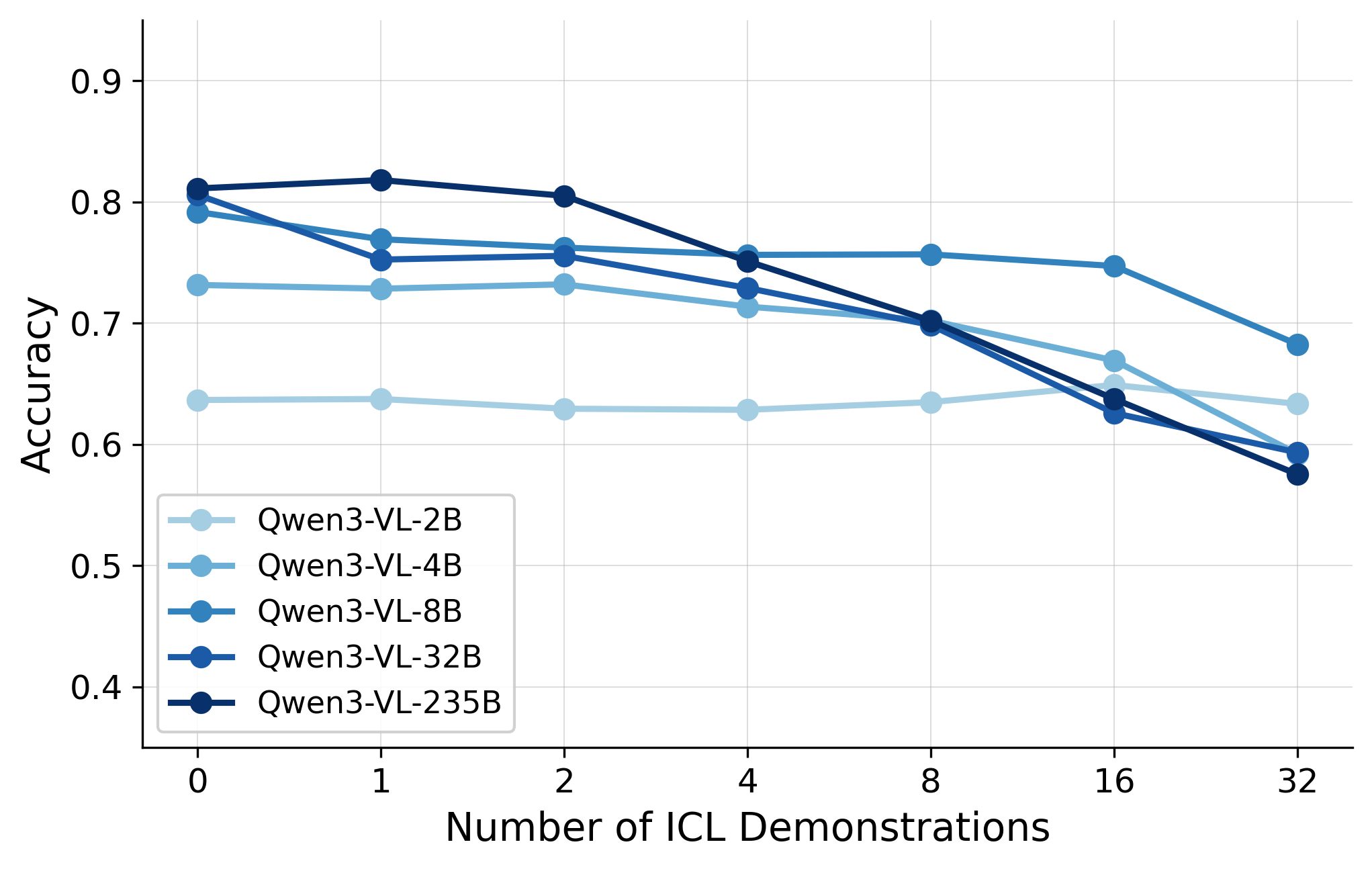}
    \caption{\textbf{Qwen3-VL unlearns under label shuffling.} In the \emph{simple} \setting, shuffled demonstration labels reduce performance below the zero-shot baseline, with larger drops for larger models.}
    \label{fig:qwen3vl_unlearn}
\end{figure}

Random label shuffling provides a further content control used in prior ICL work~\citep{min2022rethinking, kossen2024incontextlearninglearnslabel, wei2023largerlanguagemodelsincontext, pan2023incontextlearninglearnsincontext}.
It preserves the images and prompt format but breaks the mapping between each image and answer.
Figure~\ref{fig:qwen3vl_unlearn} shows that the resulting drop grows with model size.
At 32 shuffled demonstrations, Qwen3-VL 235B falls from $81.1\%$ to $57.5\%$, while 2B changes from $63.7\%$ to $63.4\%$.
These controls are inconsistent with an explanation based only on prompt length or format.

% Source: rebuttal/R2R.md, Tables R2.2--R2.4, plus the shuffled-label endpoints.
\begin{table*}[t]
  \centering
  \footnotesize
  \begin{subtable}[t]{0.59\textwidth}
    \centering
    \setlength{\tabcolsep}{4pt}
    \begin{tabular}{@{}lrrrr@{}}
      \toprule
      \textbf{Task} & \textbf{0-shot} & \textbf{16-shot} & \shortstack{\textbf{Ground}\\\textbf{mask}} & \textbf{Blank} \\
      \midrule
      Counting & 32.1 & 54.4 & 24.3 & 28.4 \\
      Color ID & 80.3 & 92.2 & 68.2 & 76.2 \\
      Shape ID & 86.4 & 94.5 & 73.8 & 81.5 \\
      \bottomrule
    \end{tabular}
    \caption{Grounding-mask and blank-image controls at 16 shots.}
    \label{tab:grounding_blank_controls}
  \end{subtable}
  \hfill
  \begin{subtable}[t]{0.37\textwidth}
    \centering
    \setlength{\tabcolsep}{5pt}
    \begin{tabular}{@{}lrr@{}}
      \toprule
      \textbf{Task} & \textbf{0-shot} & \shortstack{\textbf{Prior}\\\textbf{mask}} \\
      \midrule
      Counting & 68.2 & 63.5 \\
      Color ID & 95.6 & 93.2 \\
      Shape ID & 96.3 & 95.6 \\
      \bottomrule
    \end{tabular}
    \caption{Prior-cue masking on real images.}
    \label{tab:real_prior_mask_control}
  \end{subtable}

  \medskip
  \begin{subtable}[t]{0.59\textwidth}
    \centering
    \setlength{\tabcolsep}{4pt}
    \begin{tabular}{@{}lrrrrr@{}}
      \toprule
      \textbf{Task} & \textbf{0-shot} & \textbf{Full} & \textbf{Unrelated} & \shortstack{\textbf{Vision}\\\textbf{only}} & \textbf{Blank} \\
      \midrule
      Counting & 32.1 & 54.4 & 43.5 & 38.9 & 28.4 \\
      \bottomrule
    \end{tabular}
    \caption{Length-matched demonstration content at 16 shots.}
    \label{tab:length_matched_controls}
  \end{subtable}
  \hfill
  \begin{subtable}[t]{0.37\textwidth}
    \centering
    \setlength{\tabcolsep}{5pt}
    \begin{tabular}{@{}lrrr@{}}
      \toprule
      \textbf{Model} & \textbf{0-shot} & \textbf{Shuffled} & $\boldsymbol{\Delta}$ \\
      \midrule
      2B & 63.7 & 63.4 & $-0.3$ \\
      235B & 81.1 & 57.5 & $-23.6$ \\
      \bottomrule
    \end{tabular}
    \caption{Qwen3-VL endpoints with 32 shuffled demonstrations.}
    \label{tab:shuffled_label_endpoints}
  \end{subtable}
  \caption{\textbf{Masking and length-matched control ablations.}
  Panels (a), (b), and (c) report accuracy (\%) for Qwen3-VL 32B on the \emph{complex} \setting with greedy decoding.
  Ground mask removes grounding-cue tokens from the query image; Blank replaces each demonstration image with a same-size blank image.
  Unrelated uses demonstrations from another split of the same task, while Vision only removes question text from both the demonstrations and query.
  Panel (d) reports \emph{simple}-setting endpoints from the shuffled-label sweep in Figure~\ref{fig:qwen3vl_unlearn}; $\Delta$ is the change from zero-shot accuracy in percentage points.
  Compute limits restrict panel (c) to counting.}
  \label{tab:control_ablations}
\end{table*}

% Qwen3-VL unlearning under shuffled labels.

\paragraph{Cross-Task Demonstrations.}
Table~\ref{tab:cross_task_demonstrations} extends the counting controls to color and shape identification demonstrations.
Animal-color demonstrations improve accuracy by $2.5$ percentage points over zero shot, while the other cross-task conditions reduce accuracy.
Same-task counting demonstrations transfer better across image domains, suggesting that added context alone does not explain the gains.

% Source: rebuttal/R3D.md, demonstration table and sampling description.
% Model attribution follows the controls extended here: rebuttal/R3R.md,
% Section 4, Table R3.1 (also rebuttal/R2R.md, Table R2.4).
% R3D does not restate the model or decoding for the new cross-task rows;
% greedy decoding is explicit only for the original controls in R3R/R2R.
\begin{table}[!htbp]
  \centering
  \caption{\textbf{Cross-task demonstrations.} Accuracy on \emph{complex} counting queries for Qwen3-VL 32B. Mixed color demonstrations are sampled uniformly across splits.}
  \label{tab:cross_task_demonstrations}
  \small
  \setlength{\tabcolsep}{8pt}
  \begin{tabular}{@{}llrr@{}}
    \toprule
    \textbf{Demonstration Task} & \textbf{Images} & \textbf{Shots} & \textbf{Accuracy (\%)} \\
    \midrule
    None     & ---                & 0  & 32.1 \\
    Counting & Complex (standard) & 16 & 54.4 \\
    Counting & 2D shapes          & 16 & 43.5 \\
    \midrule
    Color ID & Animals            & 16 & 34.6 \\
    Color ID & 2D shapes          & 16 & 20.2 \\
    Color ID & Mixed splits       & 16 & 27.1 \\
    Shape ID & Traffic signs      & 16 & 25.2 \\
    \bottomrule
  \end{tabular}
\end{table}

\subsubsection{Explicit Language Prior: Can ICL Overcome?}
\label{app:explicit_prior_results}
\label{app:explicit_prior}

We next test whether ICL can override an \emph{explicitly stated} language prior, following the setup of \citet{luo2025probingvisuallanguagepriors}.
In this variant, a natural-language distractor statement appears immediately before the visual question on the \emph{complex} \setting of the three language-prior tasks: counting, color identification, and shape identification.
The distractor templates state that \texttt{\{animal\} typically has \{expected\_bias\} legs}, \texttt{\{animal\} is typically \{expected\_bias\}}, or \texttt{\{sign\} typically has a \{expected\_bias\} shape}.
For example, a counting query becomes: ``Dogs typically have 4 legs. How many legs does this animal have? Answer with a number in curly brackets, e.g., \{4\}.''
Demonstration prompts do not include distractor statements, and scoring follows the standard setup in Appendix~\ref{app:output_scoring}.

The counting case gives a compact stress test because the explicit sentence states the common animal-leg prior while the image may show a counterfactual leg count.
Table~\ref{tab:explicit_lp} compares standard counting prompts with the explicit-prior variant.
The distractor lowers zero-shot accuracy from $33.0\%$ to $28.9\%$, showing that the textual cue makes the prior harder to ignore.
With 16 demonstrations, accuracy still rises by $22.1\%$ under the explicit prior, close to the $20.7\%$ gain on the original prompts.

% Compact counting result for the explicit language-prior stress test.
\begin{table}[h]
    \centering
    \footnotesize
    \setlength{\tabcolsep}{8pt}
    \renewcommand{\arraystretch}{1.12}
    \begin{tabular}{lcc}
        \toprule
        \textbf{Counting} & \textbf{Original} & \textbf{Explicit Language Prior} \\
        \midrule
        0-shot & 33.0 & 28.9 \\
        16-shot & 53.7 & 51.0 \\
        $\Delta$ & $+20.7$ & $+22.1$ \\
        \bottomrule
    \end{tabular}
    \vspace{0.35em}
    \caption{\textbf{Counting under an explicit language prior.} Accuracy (\%) for standard counting prompts and prompts prepended with an explicit prior statement. The explicit prior lowers zero-shot accuracy, but the 0-to-16-shot gain remains comparable to the original prompt.}
    \label{tab:explicit_lp}
\end{table}

\subsubsection{Does Reasoning Help?}
\label{subsec:reasoning_icl}

We test whether reasoning improves in-context learning by enabling each model's thinking mode, which lets it generate reasoning tokens before producing an answer. 
We compare Qwen3-VL's Instruct and Thinking variants, and we toggle thinking mode on and off for Gemma-4. 
Reasoning runs use a 1024-token budget; non-reasoning runs use 512. 
Following~\citet{vo2025visionlanguagemodelsbiased}, we expect textual reasoning to offer little benefit for visual tasks.

\begin{figure}[t]
    \centering
    \includegraphics[width=\linewidth]{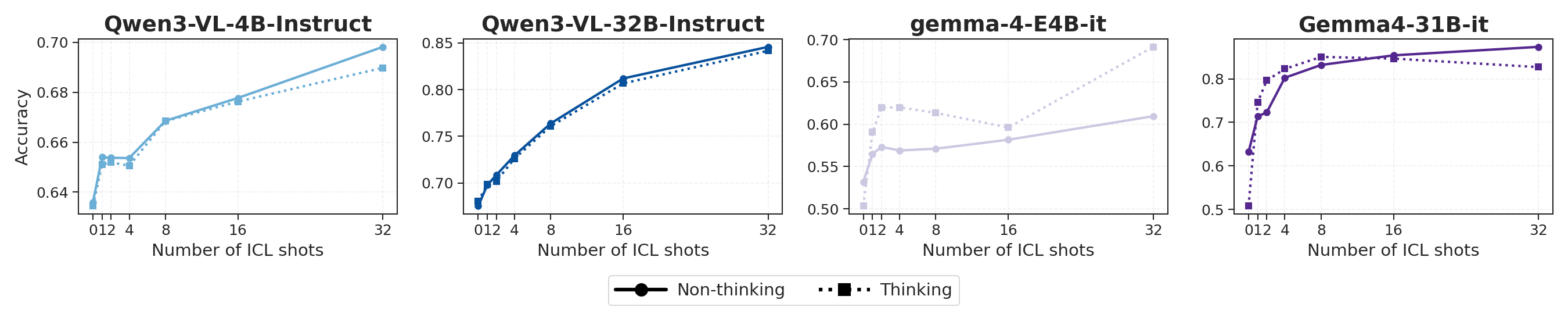}
    \caption{\textbf{Reasoning (thinking) vs.\ non-reasoning ICL accuracy.} Overall accuracy for Qwen3-VL-4B (Instruct vs. Reasoning), Qwen3-VL-32B (Instruct vs. Reasoning), Gemma-4-E4B-it, and Gemma-4-31B-it. Solid lines denote non-thinking mode; dotted lines denote thinking mode. Enabling thinking does not consistently improve ICL scaling.}
    \label{fig:app_reasoning_accuracy}
\end{figure}

Figure~\ref{fig:app_reasoning_accuracy} reports accuracy for both models at small and large scales. 
For Qwen3-VL, the Thinking variant matches the Instruct variant at both scales. 
For Gemma-4, the picture is split: E4B improves with thinking enabled, while 31B shows no consistent gain.

% \paragraph{Example Reasoning Trace.}
% \TODO{Add a representative reasoning trace comparing a correct visually grounded answer with an error that follows the prior/default answer.}

\subsubsection{Probability of Correct Answer Labels}
\label{app:answer_label_probability}

To gain deeper insight into model behavior, we examine the probability assigned to each ground-truth answer label. Specifically, we extract the output logit at the answer position and compute the probability of the correct token. For visualization, we focus on the \emph{complex} \settings of color identification and counting, which best illustrate how the model's output distribution shifts as in-context demonstrations are provided.

\begin{figure}[t]
    \centering
    \includegraphics[width=\linewidth]{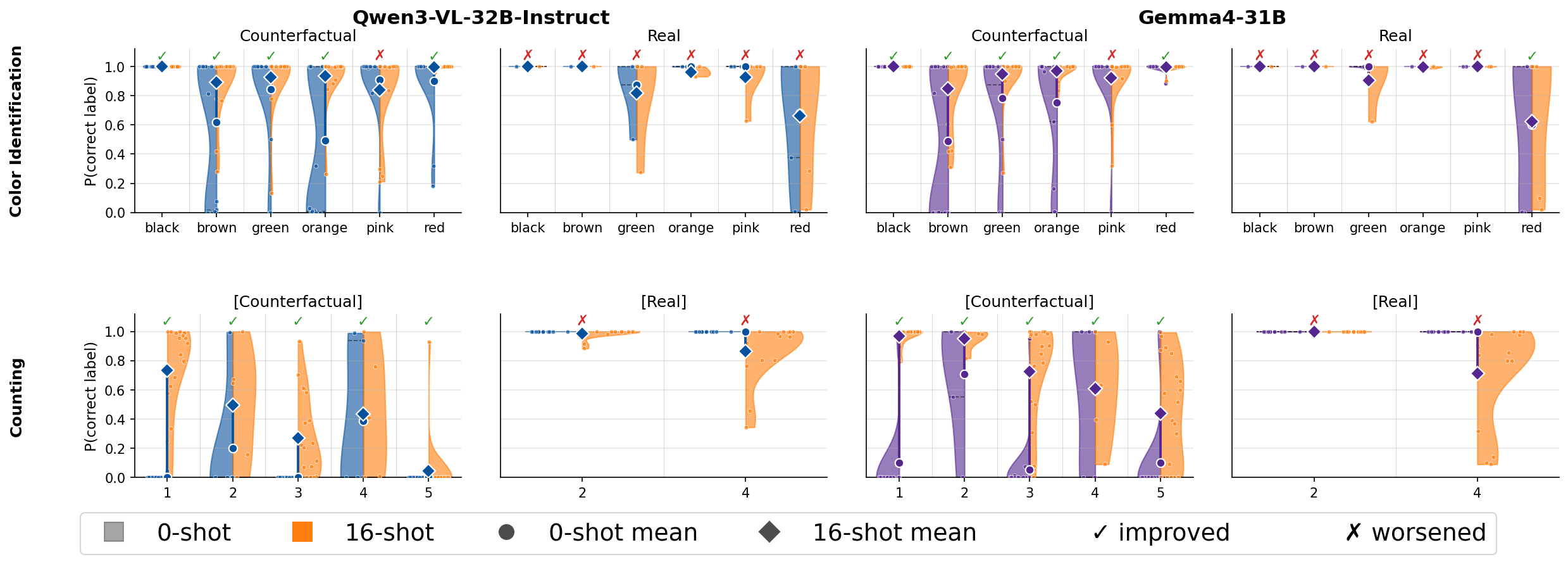}
    \caption{\textbf{Answer-label distribution shifts under ICL in the \emph{complex} \setting.} For Qwen3-VL-32B-Instruct and Gemma4-31B, we plot the probability assigned to the correct label at 0-shot and 16-shot for color identification (top row) and counting (bottom row), separated into counterfactual and real entries. Probability mass on counterfactual answers increases substantially with demonstrations, while probabilities on real entries remain largely intact.}
    \label{fig:app_answer_labels}
\end{figure}

Figure~\ref{fig:app_answer_labels} shows how the answer probabilities evolve for Qwen3-VL 32B and Gemma-4 31B. Across both models, the probability of the correct answer on counterfactual images improves dramatically with additional demonstrations, while the probability on real images remains largely unchanged. Two patterns stand out:

\begin{itemize}
    \item \textbf{Color identification (top row):} The correct-answer probability becomes sharply concentrated near 1.0, indicating that in-context demonstrations help the model overcome occasional errors induced by the \prior.
    \item \textbf{Counting:} At zero-shot, many correct answers that conflict with world knowledge encoded in the \prior receive a near-zero probability. The substantial improvement from 0-shot to 16-shot demonstrates the model's ability to override its strong \prior and learn the visual task from in-context demonstrations.
\end{itemize}

\subsection{Mechanistic Experiments}
\label{app:mechanistic_details}

This section gives full results and implementation details for the mechanistic analyses in Section~\ref{sec:mechanistic_evaluation}.
The attention analyses use the \emph{complex} \settings of counting and color identification, where each example has a visually correct answer and a competing prior answer.
We report attention shifts for Qwen3-VL 4B and 32B and Gemma-4 E4B and 31B.
Prior-cue masking covers all five tasks in the \emph{complex} \setting for Qwen3-VL 32B and Gemma-4 31B.

\subsubsection{Results Across Model Families}
\label{app:cross_family_mechanistic}

Table~\ref{tab:cross_family_mechanistic}(a) reports the change in attention per token from 0 to 16 shots.
All four models reduce attention to prior cues and increase attention to grounding cues.
For Gemma-4 E4B, these changes are $-2.62$ and $+0.71$ in units of $10^{-3}$; for Gemma-4 31B, they are $-2.02$ and $+1.23$.
The comparison concerns the direction of each model's attention shift; per-token magnitudes also depend on the model's visual-token count.

Table~\ref{tab:cross_family_mechanistic}(b) gives the full masking accuracies behind Figure~\ref{fig:4_2_qualitative}(b), alongside the 16-shot results.
Prior-cue masking exceeds both zero-shot accuracy and the matched random-mask control on every task for both models.
Gemma-4 31B improves most on counting, from $40.9\%$ to $58.8\%$, while 16-shot prompting reaches $73.9\%$.
Its direction and spatial-relation gains under prior-cue masking are smaller: $2.6$ and $1.7$ percentage points, respectively.
On spatial relation, prior-cue masking reaches $74.2\%$, compared with $73.4\%$ at 16 shots and $72.5\%$ at zero shot.
Masking and demonstrations therefore improve accuracy in the same direction, but their relative gains vary by task.

% Sources: rebuttal/R2D.md, Tables 1--2; rebuttal/R2R.md, Table R2.1.
\begin{table}[!htbp]
  \centering
  \small
  \renewcommand{\arraystretch}{1.12}
  \begin{subtable}[t]{0.94\linewidth}
    \centering
    \setlength{\tabcolsep}{6pt}
    \begin{tabular*}{\linewidth}{@{}l@{\extracolsep{\fill}}rrr@{}}
      \toprule
      \textbf{Model} & \textbf{Prior Cues} & \textbf{Grounding Cues} & \textbf{Other Tokens} \\
      \midrule
      Qwen3-VL 4B & $-3.41$ & $+0.59$ & $+0.20$ \\
      Qwen3-VL 32B & $-2.91$ & $+1.21$ & $+0.18$ \\
      Gemma-4 E4B & $-2.62$ & $+0.71$ & $+0.21$ \\
      Gemma-4 31B & $-2.02$ & $+1.23$ & $+0.16$ \\
      \bottomrule
    \end{tabular*}
    \caption{Change in attention per token from 0 to 16 shots ($\times 10^{-3}$).}
  \end{subtable}

  \medskip
  \begin{subtable}[t]{0.94\linewidth}
    \centering
    \setlength{\tabcolsep}{6pt}
    \begin{tabular*}{\linewidth}{@{}l@{\extracolsep{\fill}}rrrr@{}}
      \toprule
      \textbf{Task} & \textbf{0-shot} & \textbf{Random Mask} & \textbf{Prior Mask} & \textbf{16-shot} \\
      \midrule
      \multicolumn{5}{@{}l}{\textit{Qwen3-VL 32B}} \\
      Counting & 32.1\% & 28.2\%~{\scriptsize($-3.9$)} & 46.3\%~{\scriptsize($+14.2$)} & \textbf{54.4\%}~{\scriptsize($+22.3$)} \\
      Color ID & 80.3\% & 83.8\%~{\scriptsize($+3.5$)} & \textbf{95.3\%}~{\scriptsize($+15.0$)} & 92.2\%~{\scriptsize($+11.9$)} \\
      Shape ID & 86.4\% & 79.2\%~{\scriptsize($-7.2$)} & 92.6\%~{\scriptsize($+6.2$)} & \textbf{94.5\%}~{\scriptsize($+8.1$)} \\
      Object Direction & 73.2\% & 70.0\%~{\scriptsize($-3.2$)} & 79.3\%~{\scriptsize($+6.1$)} & \textbf{82.9\%}~{\scriptsize($+9.7$)} \\
      Spatial Relation & 64.8\% & 65.3\%~{\scriptsize($+0.5$)} & 68.2\%~{\scriptsize($+3.4$)} & \textbf{71.0\%}~{\scriptsize($+6.2$)} \\
      \midrule
      \multicolumn{5}{@{}l}{\textit{Gemma-4 31B}} \\
      Counting & 40.9\% & 36.5\%~{\scriptsize($-4.4$)} & 58.8\%~{\scriptsize($+17.9$)} & \textbf{73.9\%}~{\scriptsize($+33.0$)} \\
      Color ID & 82.5\% & 82.5\%~{\scriptsize($0.0$)} & 91.7\%~{\scriptsize($+9.2$)} & \textbf{92.8\%}~{\scriptsize($+10.3$)} \\
      Shape ID & 92.6\% & 92.1\%~{\scriptsize($-0.5$)} & 95.8\%~{\scriptsize($+3.2$)} & \textbf{99.1\%}~{\scriptsize($+6.5$)} \\
      Object Direction & 76.7\% & 70.0\%~{\scriptsize($-6.7$)} & 79.3\%~{\scriptsize($+2.6$)} & \textbf{87.8\%}~{\scriptsize($+11.1$)} \\
      Spatial Relation & 72.5\% & 70.6\%~{\scriptsize($-1.9$)} & \textbf{74.2\%}~{\scriptsize($+1.7$)} & 73.4\%~{\scriptsize($+0.9$)} \\
      \bottomrule
    \end{tabular*}
    \caption{Complex-setting accuracy (\%) with greedy decoding; best per row in bold.
    Parentheses show percentage-point changes from each model's 0-shot baseline.}
  \end{subtable}
  \caption{\textbf{Mechanistic results across Qwen3-VL and Gemma-4.}
  (a) Changes in answer-token attention to prior cues, grounding cues, and other query-image tokens.
  Attention is normalized over query-image visual tokens and averaged per token within each group.
  (b) Zero-shot accuracy, zero-shot masking interventions, and 16-shot accuracy on all five tasks.
  The random mask matches the prior-cue mask's token budget.
  Object direction and spatial relation use object identity as a proxy prior cue.}
  \label{tab:cross_family_mechanistic}
\end{table}

\subsubsection{Additional Setting Details}

\paragraph{Cue-Token Metadata Fields.}
For each example, we instantiate cue descriptors and scoring labels from three metadata fields: \texttt{prior\_cues}, \texttt{grounding\_cues}, and \texttt{ground\_truth}.
The main analyses use two cue types: prior cues from \texttt{prior\_cues} and grounding cues from \texttt{grounding\_cues}.
The \texttt{ground\_truth} field defines the visually correct answer used for scoring.
It is not, by itself, the prior-cue descriptor; it overlaps with the grounding descriptor only when the answer word also names the visible evidence, as in color identification.

For example, in a color-identification query with a red tiger, \texttt{prior\_cues} contains \texttt{tiger}, \texttt{grounding\_cues} contains the visible color \texttt{red}, and \texttt{ground\_truth} is \texttt{red}.
The prior cue is therefore the animal identity, while the grounding cue is the visible color evidence.
In a counting query with a three-legged peacock, \texttt{prior\_cues} contains \texttt{peacock}, \texttt{grounding\_cues} contains words such as \texttt{legs}, and \texttt{ground\_truth} is \texttt{3}.
The prior cue is again the animal identity, while the grounding cue is the visible leg evidence needed to count.

For object direction and spatial relation, \texttt{prior\_cues} contains the object named in the caption.
These tasks have no single prior answer; object identity serves as a proxy cue that may correlate with the answer without specifying a direct shortcut.
Masking therefore targets visual tokens that decode to this object identity.

Figure~\ref{fig:app_token_mask_exp} shows representative examples of these fields in practice: the left panel is a counting example (one-legged swan, prior cue \texttt{swan}, grounding cue \texttt{legs}), and the right panel is a color-identification example (pink raven, prior cue \texttt{raven}, grounding cue \texttt{pink}), with prior-cue tokens marked by a red overlay on the query image.

\paragraph{Logit Lens and LatentLens.}
We use logit-lens~\citep{nostalgebraist2020logitlens} and LatentLens~\citep{krojer2026latentlensrevealinghighlyinterpretable} probes to assign approximate semantic labels to visual-token representations.
The probes are used only for analysis: they identify candidate prior-cue and grounding-cue visual tokens from the model's internal representations, but they do not provide supervision to the model during generation.

\paragraph{Adaptive thresholding with Kneedle.}
\label{sec:kneedle}

Logit lens and LatentLens do not assign a single label to each visual token; instead, they produce a ranked list of text tokens that are semantically compatible with the visual representation. 
Because score scales vary across layers, probes, models, and images, we avoid using a fixed similarity threshold. For each visual token, we retrieve the top-$K$ matched text tokens, with $K=50$, sort them by decreasing score, and apply the Kneedle algorithm~\citep{ANTUNES2025102161Kneedle} to the normalized score curve.

Kneedle identifies the rank at which the score curve changes from a steep drop to a flatter tail. 
We use this knee as an adaptive cutoff: all text tokens before the knee are preserved as the decoded semantic set for that visual token. 
If no stable knee is detected, we fall back to a fixed conservative cutoff and retain all the top-$50$ tokens.

We then mark a visual token as a prior cue if any preserved text token
matches the prior-cue vocabulary, and as a grounding cue if any
preserved text token matches the grounding-cue vocabulary. A single
visual token can in principle satisfy both criteria; in such cases we
assign the token to the prior-cue set. The prior-cue and grounding-cue
sets are therefore disjoint by construction. This tie-breaking rule is
conservative for the masking analysis in
Sec.~\ref{sec:mechanistic_evaluation}: when a token simultaneously
decodes to prior-favoring and grounding semantics, the prior-cue mask
removes it, ablating any grounding evidence it also carries.
Improvements from prior-cue masking thus \emph{lower-bound} the
contribution of prior-cue semantics rather than overstate it. 

% Source: rebuttal/R2R.md, Section 5. Raw probe outputs and per-layer
% distributions are unavailable. The cue-size report does not state its cutoff.
\paragraph{Cue Sizes and Cutoff Sensitivity.}
\label{app:cue_diagnostics}
On 200 \emph{complex} counting images, $12.3\%$ of visual tokens match prior cues and $8.4\%$ match grounding cues in at least one probed early, middle, or late layer.
These are across-layer coverage rates over Qwen3-VL's query-image visual tokens.
With Kneedle disabled, retaining all $K=50$ text matches, only 32 of approximately 350K visual tokens overlap before tie-breaking across the tested tasks and datasets.
This checks overlap under the largest retained semantic sets, rather than the sensitivity of attention or masking effects to the cutoff.

\paragraph{Average Attention per Visual Token.}
\label{app:mechanistic_attention}

For each query image, we identify cue-token sets using the metadata descriptors defined above.
Let $\mathcal{V}_q$ denote all visual tokens for the query image, and let $\mathcal{C} \subseteq \mathcal{V}_q$ denote a cue-token set.
We examine attention across early, middle, and late layers.
For a model with $L$ language-model layers, the early set contains the first three layers, the middle set contains the three layers centered at $\lfloor L/2 \rfloor$, and the late set contains the last three layers.
For a selected layer $\ell$ and answer position $a$, we average attention from $a$ to each query-image visual token across heads, giving $\bar{A}^{(\ell)}_{a,i}$ for token $i$.
We compute the mean normalized attention per visual token for layer $\ell$ as
\begin{equation}
    \alpha_{\ell}(\mathcal{C}_{\ell})
    =
    \frac{1}{|\mathcal{C}_{\ell}|}
    \sum_{i \in \mathcal{C}_{\ell}}
    \frac{\bar{A}^{(\ell)}_{a,i}}
    {\sum_{j \in \mathcal{V}_q} \bar{A}^{(\ell)}_{a,j}}.
\end{equation}
The denominator restricts the normalization to attention mass assigned to query-image visual tokens.
The factor $1/|\mathcal{C}_{\ell}|$ makes cue sets of different sizes comparable.
Thus $\alpha_{\ell}(\mathcal{C}_{\ell})$ measures how much query-image attention an average token in cue set $\mathcal{C}_{\ell}$ receives at layer $\ell$.
We report the aggregate mean over the selected early, middle, and late layers:
\begin{equation}
    \alpha(\mathcal{C})
    =
    \frac{1}{|\mathcal{L}_{\mathcal{C}}|}
    \sum_{\ell \in \mathcal{L}_{\mathcal{C}}}
    \alpha_{\ell}(\mathcal{C}_{\ell}),
\end{equation}
where $\mathcal{L}_{\mathcal{C}}$ is the set of selected layers in which the corresponding cue set $\mathcal{C}_{\ell}$ is non-empty.
Not all selected layers contain all cue types.
When a prior or grounding cue set is empty in a layer, we omit that layer from the aggregate for that cue type rather than treating the missing cue as zero attention.

Qwen3-VL encodes images with a Vision Transformer using $16 \times 16$ patches, followed by a $2 \times 2$ spatial merge before the visual tokens are passed to the language model~\citep{bai2025qwen3vltechnicalreport}.
All images are resized to $512 \times 512$ pixels.
This produces $32 \times 32$ raw patches and $16 \times 16=256$ merged visual tokens per image.
Uniform attention over query-image visual tokens therefore corresponds to $1/256 \approx 0.0039$ attention per token.

\subsubsection{Prior-Cue Masking Example}
\label{app:mechanistic_masking}

Token ablation blocks generated answer tokens from attending to selected query-image visual tokens during generation, but leaves these tokens unmasked during prefill. 
This distinction matters because logit-lens and LatentLens probing identify cue tokens from the unmasked visual-token representations; masking during prefill would alter the very representations used for cue identification. 
Generation-time masking thus holds the probed representation fixed while isolating the effect of selected cue tokens on answer generation.

The main intervention masks prior-cue tokens identified from the \texttt{prior\_cues} field.
We report accuracy for all five tasks using Qwen3-VL 32B and Gemma-4 31B with greedy decoding (Table~\ref{tab:cross_family_mechanistic}(b)).
Where a prior (default) answer is defined, we also measure how the candidate-normalized probability mass shifts between that answer and \texttt{ground\_truth}.
A matched random-token ablation accompanies this intervention as a control: for each query, we mask the same number of visual tokens as in the prior-cue mask for that query, drawn uniformly from visual tokens that are flagged as neither prior cues nor grounding cues, and average over multiple random seeds.
Prior-cue masking outperforms this matched-budget random control on all five tasks in both models (Figure~\ref{fig:4_2_qualitative}(b)).
The object-identity proxy makes the evidence less direct for direction and relation, where Gemma's gains are also smaller than on counting, color, and shape.

\begin{figure}[h]
    \centering
    \includegraphics[width=\linewidth]{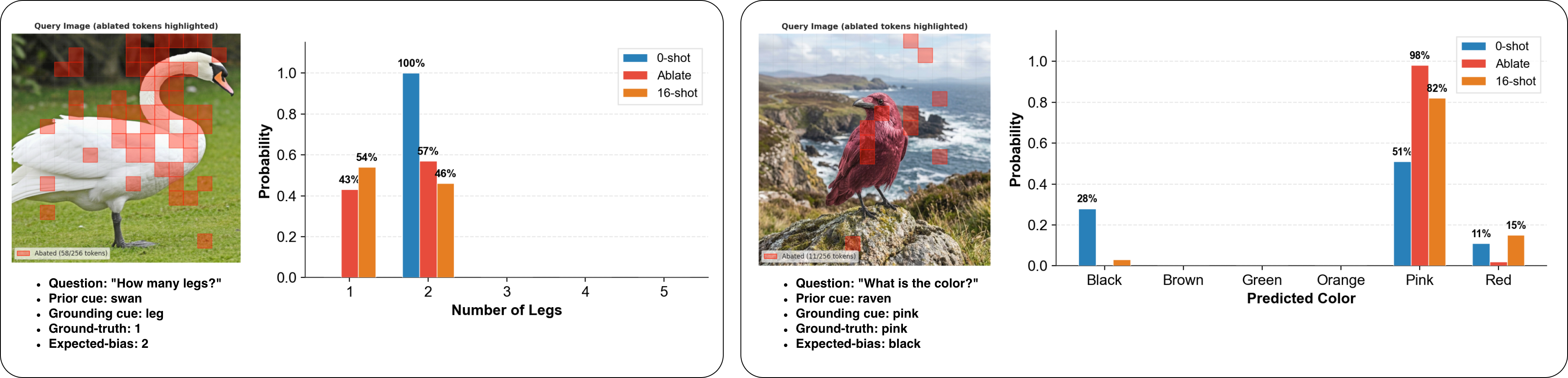}
    \caption{\textbf{Prior-cue masking examples.} Two queries from the \emph{complex} \setting with their identified prior-cue tokens (red overlay on the query image) and the candidate-normalized answer distributions at 0-shot, after prior-cue token ablation, and at 16-shot. Left: a one-legged swan, where the 0-shot model places nearly all mass on the prior-default answer ``2''; ablating the prior-cue tokens shifts mass onto the ground-truth ``1''; and 16-shot demonstrations produce a similar shift. Right: a pink raven, where 0-shot mass concentrates on the \prior default ``black'', and both prior-cue ablation and 16-shot ICL move probability mass onto the ground-truth ``pink''. Prior-cue ablation reproduces the qualitative effect of in-context demonstrations.}
    \label{fig:app_token_mask_exp}
\end{figure}

Figure~\ref{fig:app_token_mask_exp} shows two prior-cue masking examples from the \emph{complex} \setting. 
The left panel presents a one-legged swan from the counting task, where the prior cue is ``swan'' and the grounding cue is ``leg.'' 
The answer-token probabilities show that both token masking and demonstrations shift the model away from the expected-bias answer (2) toward the ground-truth answer (1). The right panel presents a pink raven from the color-identification task, where the same pattern holds: both token masking and demonstrations shift the answer toward the ground-truth answer ``pink.''

\subsection{License and Ethics Statement}
\label{app:license_ethics}

Our benchmark contains three categories of images: (1)~procedurally rendered synthetic images, (2)~real images regenerated with Gemini from base images in existing publicly available benchmarks, and (3)~images generated from scratch or counterfactually edited with Gemini. Synthetic and generated images carry no privacy or consent concerns. The licenses and terms of the source benchmarks govern the base images used for regeneration; we cite each source dataset (Appendix~\ref{app:data_construction}) and do not redistribute the original base images.

All generated and counterfactually edited images are disclosed as such in the release metadata. Each sample includes a source-type field indicating whether the image is procedural, regenerated, generated, or edited.

Our benchmark is designed to evaluate shortcut and prior behavior in VLMs. By construction, examples in the \emph{complex} \setting intentionally contain unusual or counter-stereotypical object attributes (e.g., a pink tiger, a one-legged dog). These examples do not represent real-world distributions and are not intended to reinforce stereotypical associations.

Human validation was an in-house dataset quality review by two authors.
They independently selected answers while blind to the intended labels and applied the fixed quality criteria in Appendix~\ref{app:human_validation}.
We discarded images with label disagreements or quality failures; no crowd-sourced labeling or human-subject interaction was conducted.

\paragraph{Broader Impacts.}
On the positive side, our work provides a controlled diagnostic that separates \prior dominance, where the model has the relevant visual evidence but defaults to a learned expectation, from genuine capability absence. This distinction tells researchers and practitioners when a failure can be addressed at test time with demonstrations and when it instead calls for changes to data, training, or architecture, supporting more targeted and resource-efficient VLM development. 
This distinction is also fairness-relevant: a VLM that does not actually consult the image must substitute statistical priors for perception, and those priors can encode demographic and stereotypical biases, detrimental to downstream applications.
Finally, ICL gains require long multimodal contexts, which increase inference compute and energy use (Section~5).

\paragraph{AI Use Statement.}
Generative AI tools assisted with data generation, coding the ICL pipeline, and editing the LaTeX source.
The authors reviewed the resulting source and take responsibility for the final content of this work, including text, claims, and artifacts produced with the aid of generative AI.

% NeurIPS checklist
\ifdefined\arxivversion
\else
  \newpage
  \section*{NeurIPS Paper Checklist}

\begin{enumerate}

\item {\bf Claims}
    \item[] Question: Do the main claims made in the abstract and introduction accurately reflect the paper's contributions and scope?
    \item[] Answer: \answerYes{}
    \item[] Justification: The abstract and introduction (Section~1) state our main claims---that visual ICL substantially reduces \prior dominance on prior-conflicting examples and that it does not create capabilities that are absent---and these claims are supported by the empirical results in Section~\ref{sec:performance_evaluation}, the mechanistic analyses in Section~\ref{sec:mechanistic_evaluation}, and the capability-limit analysis in Section~\ref{sec:limitations}.

\item {\bf Limitations}
    \item[] Question: Does the paper discuss the limitations of the work performed by the authors?
    \item[] Answer: \answerYes{}
    \item[] Justification: Section~\ref{sec:limitations} analyzes when ICL fails to compensate for weak underlying visual capabilities (e.g., high-cardinality counting), and the conclusion (Section~5) discusses the compute and context cost of long multimodal demonstrations along with the resulting need for demonstration selection or context distillation.

\item {\bf Theory assumptions and proofs}
    \item[] Question: For each theoretical result, does the paper provide the full set of assumptions and a complete (and correct) proof?
    \item[] Answer: \answerNA{}
    \item[] Justification: Section~\ref{sec:formal_definition} introduces a conceptual framework (training prior, task decomposition, ICL as Bayesian inference over prediction strategies) that motivates the empirical study, but the paper does not state or prove formal theorems.

    \item {\bf Experimental result reproducibility}
    \item[] Question: Does the paper fully disclose all the information needed to reproduce the main experimental results of the paper to the extent that it affects the main claims and/or conclusions of the paper (regardless of whether the code and data are provided or not)?
    \item[] Answer: \answerYes{}
    \item[] Justification: Appendix~\ref{app:data_construction} documents the data construction pipeline, prompts, and human-validation criteria for every task and \setting; Appendix~\ref{app:evaluation_settings} specifies the prompt format, demonstration sampling, contexts, and shot counts; Appendix~\ref{app:model_settings} lists exact model checkpoints and inference backends; Appendix~\ref{app:output_scoring} describes the exact-match scoring procedure.

\item {\bf Open access to data and code}
    \item[] Question: Does the paper provide open access to the data and code, with sufficient instructions to faithfully reproduce the main experimental results, as described in supplemental material?
    \item[] Answer: \answerYes{}
    \item[] Justification: We will release the benchmark (synthetic images, Gemini-regenerated, generated, and edited images, per-sample metadata with source-type fields, and task labels) together with the evaluation code. An anonymized version is included in the supplemental material at submission time.

\item {\bf Experimental setting/details}
    \item[] Question: Does the paper specify all the training and test details (e.g., data splits, hyperparameters, how they were chosen, type of optimizer) necessary to understand the results?
    \item[] Answer: \answerYes{}
    \item[] Justification: Our evaluation is training-free; all evaluation details---shot counts $N \in \{0,1,2,4,8,16,32\}$, label-balanced demonstration sampling, the 10 fixed demonstration contexts shared across queries, image resolution, decoding parameters (Table~\ref{tab:app_sampling_params}), and per-family model checkpoints---are documented in Appendix~\ref{app:evaluation_settings} and Appendix~\ref{app:model_settings}.

\item {\bf Experiment statistical significance}
    \item[] Question: Does the paper report error bars suitably and correctly defined or other appropriate information about the statistical significance of the experiments?
    \item[] Answer: \answerYes{}
    \item[] Justification: Each non-zero-shot accuracy is averaged over 10 fixed demonstration contexts, shared across queries and sampled with a fixed seed (Appendix~\ref{app:evaluation_settings}); error bars in the ICL-scaling figures are standard deviations across these contexts. Behavioral runs use stochastic decoding (Table~\ref{tab:app_sampling_params}) with one sampled output per query-context pair, so the error bars also include decoding noise and therefore upper-bound the variability due to the choice of demonstration set.

\item {\bf Experiments compute resources}
    \item[] Question: For each experiment, does the paper provide sufficient information on the computer resources (type of compute workers, memory, time of execution) needed to reproduce the experiments?
    \item[] Answer: \answerYes{}
    \item[] Justification: Appendix~\ref{app:compute} reports the compute setup: a single node with 4 NVIDIA H200 GPUs, vLLM for behavioral evaluation and Hugging Face Transformers for mechanistic analyses, with the full empirical ICL sweep consuming approximately 400 GPU-hours.
    
\item {\bf Code of ethics}
    \item[] Question: Does the research conducted in the paper conform, in every respect, with the NeurIPS Code of Ethics \url{https://neurips.cc/public/EthicsGuidelines}?
    \item[] Answer: \answerYes{}
    \item[] Justification: We have reviewed the NeurIPS Code of Ethics. The benchmark uses procedurally generated images and images generated or edited with a public image-generation API, including images regenerated from base images in existing public benchmarks; no human-subject data is collected and no personally identifying information is included (Appendix~\ref{app:license_ethics}).

\item {\bf Broader impacts}
    \item[] Question: Does the paper discuss both potential positive societal impacts and negative societal impacts of the work performed?
    \item[] Answer: \answerYes{}
    \item[] Justification: Appendix~\ref{app:license_ethics} discusses both the positive impact---a diagnostic that distinguishes prior dominance (test-time correctable) from capability absence and informs whether test-time intervention or further training is appropriate---and the steps taken to avoid reinforcing stereotypical associations from the counter-stereotypical \emph{complex} stimuli, which are explicitly flagged as non-representative of real-world distributions.
    
\item {\bf Safeguards}
    \item[] Question: Does the paper describe safeguards that have been put in place for responsible release of data or models that have a high risk for misuse (e.g., pre-trained language models, image generators, or scraped datasets)?
    \item[] Answer: \answerNA{}
    \item[] Justification: We release an evaluation benchmark, not a generative model. The dataset contains synthetic, regenerated, and counterfactually edited images that pose no clear misuse risk, and each sample carries a source-type field (Appendix~\ref{app:license_ethics}) so downstream users can distinguish real, generated, and edited images.

\item {\bf Licenses for existing assets}
    \item[] Question: Are the creators or original owners of assets (e.g., code, data, models), used in the paper, properly credited and are the license and terms of use explicitly mentioned and properly respected?
    \item[] Answer: \answerYes{}
    \item[] Justification: All evaluated VLMs are cited in Appendix~\ref{app:model_settings} with their corresponding technical reports and licenses, and all source benchmarks are cited in Appendix~\ref{app:data_construction}. License terms and re-distribution constraints are summarized in Appendix~\ref{app:license_ethics}; the source benchmarks' licenses and terms govern the base images used for regeneration, and we do not redistribute the original base images.

\item {\bf New assets}
    \item[] Question: Are new assets introduced in the paper well documented and is the documentation provided alongside the assets?
    \item[] Answer: \answerYes{}
    \item[] Justification: The new benchmark introduced in this paper, is documented in Section~\ref{sec:setting} and Appendix~\ref{app:data_construction}, including task definitions, construction and validation pipelines, label sets, dataset statistics (Table~\ref{tab:dataset_stats}), and the exact prompts used for image generation and counterfactual editing (Appendix~\ref{app:generation_prompts}); a datasheet and per-sample metadata accompany the release.

\item {\bf Crowdsourcing and research with human subjects}
    \item[] Question: For crowdsourcing experiments and research with human subjects, does the paper include the full text of instructions given to participants and screenshots, if applicable, as well as details about compensation (if any)?
    \item[] Answer: \answerNA{}
    \item[] Justification: Two authors independently labeled images while blind to the intended labels and applied fixed quality criteria (Appendix~\ref{app:human_validation}). We retained an image only when both selected the intended label and confirmed that it met these criteria. This was in-house dataset quality review; no crowd workers, external annotators, or human subjects participated.

\item {\bf Institutional review board (IRB) approvals or equivalent for research with human subjects}
    \item[] Question: Does the paper describe potential risks incurred by study participants, whether such risks were disclosed to the subjects, and whether Institutional Review Board (IRB) approvals (or an equivalent approval/review based on the requirements of your country or institution) were obtained?
    \item[] Answer: \answerNA{}
    \item[] Justification: The work does not involve human-subject data collection; all annotation was an in-house quality review of synthetic and image-model-generated stimuli, so no IRB review was required.

\item {\bf Declaration of LLM usage}
    \item[] Question: Does the paper describe the usage of LLMs if it is an important, original, or non-standard component of the core methods in this research? Note that if the LLM is used only for writing, editing, or formatting purposes and does \emph{not} impact the core methodology, scientific rigor, or originality of the research, declaration is not required.
    %this research?
    \item[] Answer: \answerYes{}
    \item[] Justification: We use Gemini~\citep{gemini_team_google_2023,googledeepmind_gemini31pro_modelcard_2026} as a non-standard core component for image generation, regeneration, and counterfactual editing in the \emph{complex} \settings of the benchmark; the exact prompts and per-task pipelines are documented in Appendix~\ref{app:data_construction} and Appendix~\ref{app:generation_prompts}. The VLMs that are the subject of study (Appendix~\ref{app:model_settings}) are evaluated through standard inference and are not used to develop the methodology itself.

\end{enumerate}

\fi

\end{document}